\documentclass[10pt,twocolumn,letterpaper]{article}

\usepackage{iccv}              

\definecolor{iccvblue}{rgb}{0.21,0.49,0.74}

\usepackage[pagebackref,breaklinks,colorlinks,allcolors=iccvblue]{hyperref}
\usepackage{algorithm}
\usepackage{algorithmicx,algpseudocode}
\usepackage{xcolor}

\def\0{\ensuremath{\mathbf{0}}}
\def\1{\ensuremath{\mathbf{1}}}

\def\n{\ensuremath{\mathbf{n}}}

\def\x{\ensuremath{\mathbf{x}}}
\def\y{\ensuremath{\mathbf{y}}}

\def\D{\ensuremath{\mathbf{D}}}

\def\F{\ensuremath{\mathbf{F}}}

\def\I{\ensuremath{\mathbf{I}}}

\def\K{\ensuremath{\mathbf{K}}}
\def\L{\ensuremath{\mathbf{L}}}

\def\Q{\ensuremath{\mathbf{Q}}}

\def\W{\ensuremath{\mathbf{W}}}
\def\X{\ensuremath{\mathbf{X}}}

\def\cG{\ensuremath{\mathcal{G}}}

\def\cN{\ensuremath{\mathcal{N}}}

\def\cV{\ensuremath{\mathcal{V}}}

\def\bPhi{\ensuremath{\boldsymbol{\Phi}}}
\def\bLambda{\ensuremath{\boldsymbol{\Lambda}}}

\def\ie{\textit{i.e.}}
\def\eg{\textit{e.g.}}

\newcommand{\ourmethod}{{GIGA-ToF}\xspace}

\def\paperID{3338} 
\def\confName{ICCV}
\def\confYear{2025}

\title{Consistent Time-of-Flight Depth Denoising via Graph-Informed Geometric Attention}

\author{Weida Wang$^{1*}$ \quad Changyong He$^1$\thanks{ indicates equal contribution.} \quad Jin Zeng$^{1}$\thanks{Corresponding author: Jin Zeng (zengjin@tongji.edu.cn).} \quad Di Qiu$^2$\\
$^1$School of Computer Science and Technology, Tongji University \quad $^2$Google
}

\begin{document}
\maketitle
\begin{abstract}
Depth images captured by Time-of-Flight (ToF) sensors are prone to noise, requiring denoising for reliable downstream applications.
Previous works either focus on single-frame processing, or perform multi-frame processing without considering depth variations at corresponding pixels across frames, leading to undesirable temporal inconsistency and spatial ambiguity.
In this paper, we propose a novel ToF depth denoising network leveraging motion-invariant graph fusion to simultaneously enhance temporal stability and spatial sharpness.
Specifically, despite depth shifts across frames, graph structures exhibit temporal self-similarity, enabling cross-frame geometric attention for graph fusion.
Then, by incorporating an image smoothness prior on the fused graph and data fidelity term derived from ToF noise distribution, we formulate a maximum a posterior problem for ToF denoising.
Finally, the solution is unrolled into iterative filters whose weights are adaptively learned from the graph-informed geometric attention, producing a high-performance yet interpretable network.
Experimental results demonstrate that the proposed scheme achieves state-of-the-art performance in terms of accuracy and consistency on synthetic DVToF dataset and exhibits robust generalization on the real Kinectv2 dataset.
Source code will be released at \href{https://github.com/davidweidawang/GIGA-ToF}{\textcolor{blue}{https://github.com/davidweidawang/GIGA-ToF}}.
\end{abstract}





\section{Introduction}
\label{sec:intro}



\begin{figure}[t]
\centering
\includegraphics[width=\columnwidth]{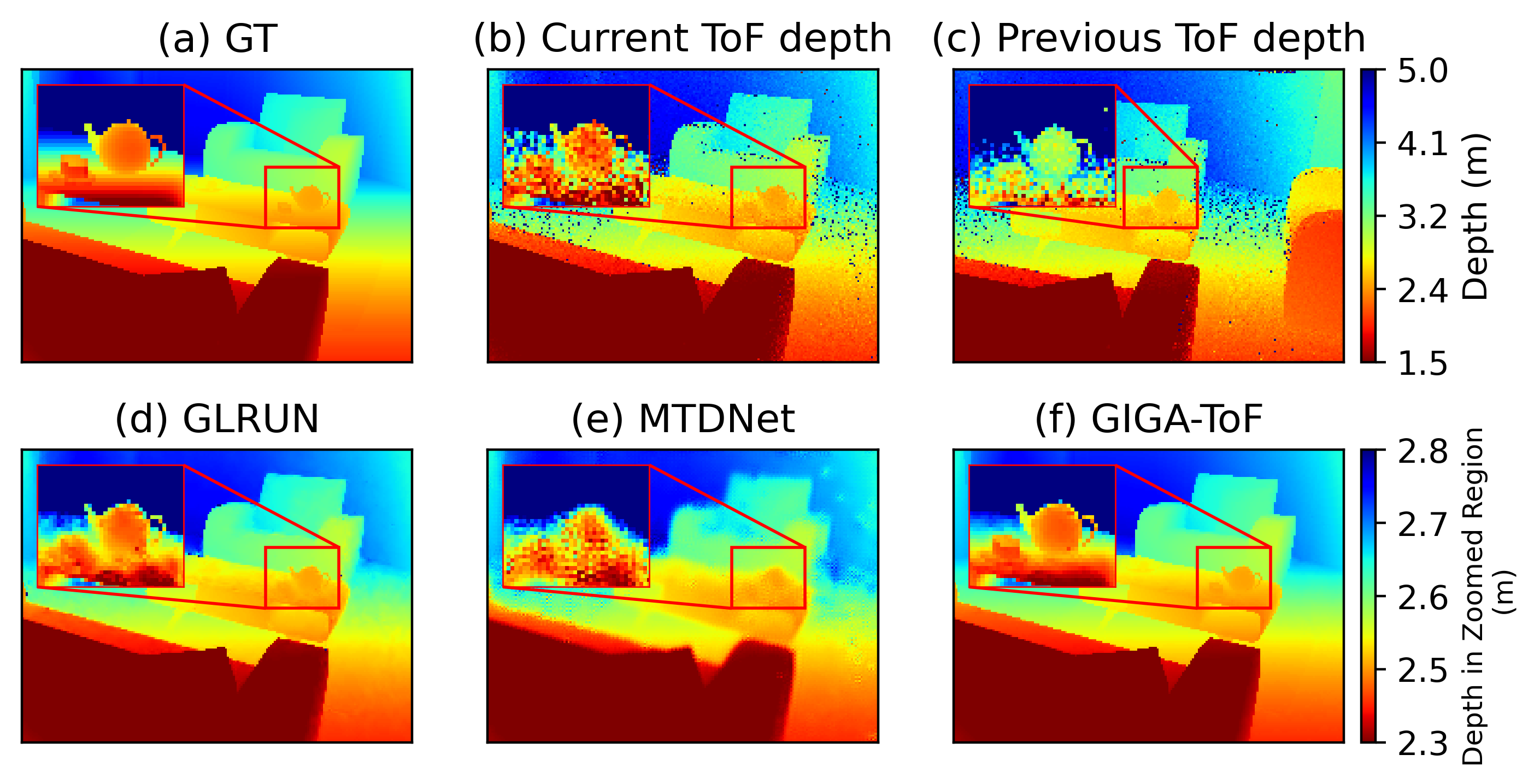}
\vspace{-0.8cm}
\caption{Illustration of (a) GT depth, noisy ToF depth in (b) current frame and (c) previous frame, (d) single-frame GLRUN \cite{jia2025deep} where noise remains, (e) multi-frame MTDNet \cite{dong2024exploiting} fusing \textit{depth features}, (f) proposed GIGA-ToF fusing \textit{graph structures}. 
Due to depth shifts at corresponding pixels in red rectangles, (e) loses details, while (f) removes noise while preserving sharpness because the neighborhood correlation graphs are motion-invariant.}
\label{fig:intro}
\vspace{-0.60cm}
\end{figure}

Continuous-wave Time-of-Flight (ToF) sensing \cite{bhandari2016signal} has emerged as the mainstream 3D imaging scheme due to its real-time response speed and low power consumption, empowering various applications such as 
robotics \cite{miki2022learning}, 3D reconstruction \cite{kang2021gradient}, augmented reality \cite{du2020depthlab}, \etc. 
For brevity, we hereinafter refer to continuous-wave ToF sensors as ToF sensors.
However, depth images captured by ToF sensors are subject to noise at distant, low-reflectance, glossy areas \cite{chen2022configurable} as shown in Fig.\,\ref{fig:intro}(b), which significantly impedes their performance in advanced applications. 

To enhance the quality of ToF depth images, researchers have proposed a variety of denoising methods. 
Early work primarily focused on statistical model-based filtering techniques, such as bilinear \cite{xiang2016libfreenect2} and non-local means \cite{frank2009denoising} filtering. 
Leveraging progress in \textit{graph signal processing} (GSP), ToF depth denoising is formulated as a \textit{maximum a posteriori} (MAP) problem using graph-based image priors to promote depth image properties such as sparsity \cite{hu2013depth} or smoothness \cite{rossi2020joint,zhang2022graph}.
With the advent of deep learning, methods based on deep neural networks (DNNs) achieve the state-of-the-art (SOTA) performance \cite{su2018deep,chen2020very,qiao2022depth,gutierrez2021itof2dtof}. 
However, most existing DNN schemes focus on \textit{single-frame processing} and ignore cross-frame correlations, resulting in undesirable \textit{temporal inconsistency}.
Fig.\,\ref{fig:intro}(d) exemplifies the single-frame scheme GLRUN \cite{jia2025deep}, where the result contains noticeable noise due to limited intra-frame information for denoising, further leading to temporal jittering. 


This motivates recent multi-frame processing methods exploiting temporal correlation inherent in ToF depth video.
These methods typically estimate scene flow \cite{sun2023consistent,li2023temporally} or inter-frame correlation \cite{dong2024exploiting} to establish correspondence between pixels in different frames, based on which the features of corresponding or correlated pixels are fused to reconstruct the final depth output.
However, depth values of the same object are changing in different frames due to camera motions as shown in Fig.\,\ref{fig:intro} by comparing (b) and (c), so features extracted from depth are usually inconsistent across frames.
Due to the \textit{temporal variation of depth}, direct fusion of depth features is likely to result in \textit{spatial ambiguity} where features with shifts are aggregated.
Fig.\,\ref{fig:intro}(e) exemplifies the multi-frame processing MTDNet \cite{dong2024exploiting} fusing depth features, resulting in loss of details.

On the contrary, we fuse \textit{motion-invariant graph structures}, simultaneously enhancing temporal stability and spatial sharpness.
As illustrated in Fig.\,\ref{fig:intro}(b) and (c), despite the depth value shifts, the graph structures reflecting correlations among neighboring pixels are similar in current and previous frames, \ie, representing the shape of the teapot.
This motivates us to construct intra-frame graphs to encode pixel correlations within depth images, then establish cross-frame geometric attention to fuse graphs in current and reference frames.
In this way, the temporal correlation is efficiently utilized to generate smooth results with spatial sharpness as shown in Fig.\,\ref{fig:intro}(f).

Apart from the spatial ambiguity issue, existing DNN schemes are usually trained on synthetic data due to the difficulty in acquiring ground truth \cite{su2018deep,guo2018tackling}, resulting in \textit{poor generalization} to real data.
Although existing schemes adopt domain adaptation to enhance the network robustness to real noise \cite{agresti2019unsupervised,agresti2022unsupervised}, the performance still fails at high noise levels.
In contrast, we incorporate the image smoothness prior defined on the fused graph into the network architecture, restricting its solution space \cite{monga2021algorithm} and enhancing generalization to real data.
Specifically, leveraging the fused graph to impose image smoothness prior and incorporating the data fidelity term based on the ToF depth noise distribution, we formulate the MAP problem for denoising ToF raw data. The solution is unrolled into iterative filters whose weights are dynamically learned from the geometric attention informed by cross-frame graph fusion.
The resulting network \textit{combines high performance with graph spectral interpretability, facilitated by the graph-informed geometric attention (GIGA) module} and is referred to as \textit{GIGA-ToF} network.
Our contributions are summarized as follows.

\begin{itemize}
\item We utilize cross-frame correlation by fusing motion-invariant graph structures, which simultaneously enhances temporal consistency and spatial sharpness;
\item We formulate the MAP problem for ToF denoising by leveraging the fuse graph to impose image smoothness prior; the network is designed by unrolling the solution into iterative filtering to enable adaptive filter weight learning from the graph-informed geometric attention; 
\item We demonstrate the enhanced accuracy and consistency of GIGA-ToF on the synthetic DVToF dataset, outperforming competing schemes by at least 37.9\% in MAE and 13.2\% in TEPE. In addition, we show strong generalization ability of GIGA-ToF to real unseen Kinectv2 data.
\end{itemize}


\section{Related Works}
\label{sec:relate}


\subsection{ToF Depth Denoising}


ToF depth denoising methods can be categorized into model-based and DNN-based approaches.
Model-based methods rely on mathematical models derived from signal priors \cite{xiang2016libfreenect2,georgiev2018time}. Recently, leveraging progress in GSP \cite{ortega2018graph,cheung2018graph}, ToF depth denoising is formulated as a MAP problem using graph-based image priors \cite{hu2013depth,rossi2020joint,zhang2022graph}. However, \textit{rule-based modeling can be suboptimal in practice due to the complicated nature of real noise.}

Recent works focus on DNN-based methods for ToF denoising, which leverage large datasets and deep learning architectures to improve noise removal. 
While many approaches directly denoise generated depth images \cite{marco2017deeptof, yan2018ddrnet}, errors accumulate during depth construction from raw ToF data, resulting in distinctive ToF depth noise distributions and posing difficulty on denoising \cite{su2018deep}.
This motivates various methods to instead process raw ToF data and build end-to-end networks to produce denoised depth images \cite{su2018deep,guo2018tackling,agresti2019unsupervised,chen2020very,schelling2022radu}. 
For example, ToFNet \cite{su2018deep} generated restored depth from raw ToF data with a multi-scale network, significantly improving imaging quality.
Despite the advancements in both model-based and DNN-based approaches, most existing ToF denoising methods operate in a frame-by-frame manner, neglecting cross-frame correlation.
This results in \textit{temporal inconsistency and hinder application of ToF depth in downstream tasks}, where temporal stability is essential for robust performance.


\subsection{Temporal ToF Depth Denoising}


In practical applications, depth restoration is typically performed on video streams rather than individual frames. 
Nevertheless, there is relatively little work focusing on utilizing temporal correlation and maintaining temporal stability for ToF depth denoising.
In model-based methods, temporal correlation is utilized in signal modeling, such as motion vector smoothness prior in the graph domain \cite{yang2014graph} and patch similarity prior based on optical flow \cite{min2011depth}, but the \textit{optimization is usually computationally heavy and is infeasible for real-time processing}.

In DNN-based methods, while ConvLSTM \cite{patil2020don} fused concatenated frames without alignment, DVSR \cite{sun2023consistent} and CODD \cite{li2023temporally} estimated scene flow for multi-frame alignment, based on which the features of corresponding pixels were fused.
MTDNet \cite{dong2024exploiting} leveraged both intra- and inter-frame correlations for multi-frame ToF denoising, guided by a confidence map to prioritize regions with strong ToF noise. 
Nevertheless, since depth at corresponding pixels varies across frames \cite{li2023temporally}, \textit{directly fusing cross-frame depth features for depth reconstruction results in loss of details} as shown in Fig.\,\ref{fig:intro}(e).
Moreover, existing temporal ToF denoising networks are \textit{purely data-driven and ignorant of ToF sensing mechanism, resulting in poor generalization to real data due to difficulty in acquiring ground truth.}

In contrast, we fuse graph structures across frames which are motion-invariant and exhibit temporal self-similarity, which resolves spatial ambiguity while promoting temporal consistency.
Moreover, we incorporate the image smoothness prior based on the fused graph, together with the data fidelity term based on ToF depth noise distribution in the network design, enhancing generalization to real data.




\subsection{Generalizable ToF Depth Denoising}
Although model-based methods \cite{yang2014graph,min2011depth} without the notion of training are robust to unseen real noise, but the optimization is computationally costly.
On the other hand, DNN-based schemes achieve SOTA performance on synthetic data but are limited in generalization to real noise.
UDA \cite{agresti2019unsupervised} adopted domain adaptation to enhance network generalization ability but failed at high noise levels.
GLRUN \cite{jia2025deep} utilized algorithm unrolling of graph Laplacian regularization \cite{pang2021graph,zeng2019deep}, resulting in a robust and efficient network.
Nevertheless, these schemes focus on single-frame processing where temporal correlations are not utilized, while we develop an interpretable network based on temporal self-similarity of graph structures inherent in ToF data, enhancing accuracy and robustness to real unseen noise.



\section{ToF Imaging Mechanism Overview}
\label{sec:pre}


To measure the depth $x_d$ of an object, the laser of the ToF sensor emits a periodic signal $s_e(t)$, which is typically modulated by a sinusoidal function with frequency $f_m$. The reflected signal $s_r(t)$, captured by the sensor, exhibits a phase shift $\phi$ relative to $s_e(t)$ after the signal travels a distance of $2x_d$ \cite{zanuttigh2016time}.
%
%
$\phi$ is then measured by computing the correlation between $s_r(t)$ and a phase shifted version of $s_e(t)$ with phase offset $\theta$, resulting in raw measurements:
\begin{align}
c_{\theta} & = \lim_{T \to \infty} \frac{1}{T} \int_{-\frac{T}{2}}^{\frac{T}{2}} s_r(t) s_e(t+ \frac{\theta}{2\pi f_m }) dt   \\
  &= \frac{\alpha}{2} \cos(\phi+\theta) + \beta, \label{eq:amp}
\end{align}
where $T$ is the exposure time, $\alpha$ is the signal amplitude, $\beta$ is the ambient light intensity.
By measuring $c_{\theta}$ for multiple phase offsets $\theta$, the raw ToF pair, \ie, in-phase $x_i$ and quadrature $x_q$ components of $\phi$, are computed as \cite{schelling2022radu},
\begin{equation} \label{eq:iq}
    x_i = \sum_{\theta} \cos(\theta) c_{\theta}, ~ x_q = \sum_{\theta} - \sin(\theta) c_{\theta},
\end{equation}
so that $\phi$ is given as $\phi = \arctan (x_q/x_i)$.
Then depth $x_d$ and amplitude $x_a$ are reconstructed from $x_i$ and $x_q$ as
\begin{equation} \label{eq:d}
x_d = \frac{c \phi }{4 \pi f_m} = \frac{c \arctan (x_q/x_i) }{4 \pi f_m},  ~ x_a= \sqrt{x_i^2+x_q^2},
\end{equation}
where $c$ is the light speed.

 
\section{Problem Formulation}
\label{sec:algo}

Given $T$ continuous frames of noisy ToF raw data $\y_i^t,\y_q^t \in \mathbb{R}^N$ in vectorized form, where $t \in [1, T]$ is the frame index, $N$ is total number of pixels, we aim to recover $T$ frames of clean $\x_i^t,\x_q^t \in \mathbb{R}^N$ which is then converted to depth map $\x_d^t \in \mathbb{R}^N $ using (\ref{eq:d}).
In this section, we first define intra-frame graph modeling for each frame of ToF raw data in Sec.\,\ref{sec:3_intra}, then propose the cross-frame graph fusion strategy to exploit graph correlation in the reference frame for current frame denoising in Sec.\,\ref{sec:3_inter}.
Finally, leveraging the fused graph to impose image smoothness prior and incorporating the data fidelity term based on the ToF depth noise distribution, we formulate the MAP optimization problem for denoising ToF raw data in Sec.\,\ref{sec:3_map}.
The solution to this problem further guides the subsequent network design.

\subsection{Intra-frame Graph Modeling}
\label{sec:3_intra}

\begin{figure}[t]
\centering
\includegraphics[width=\columnwidth]{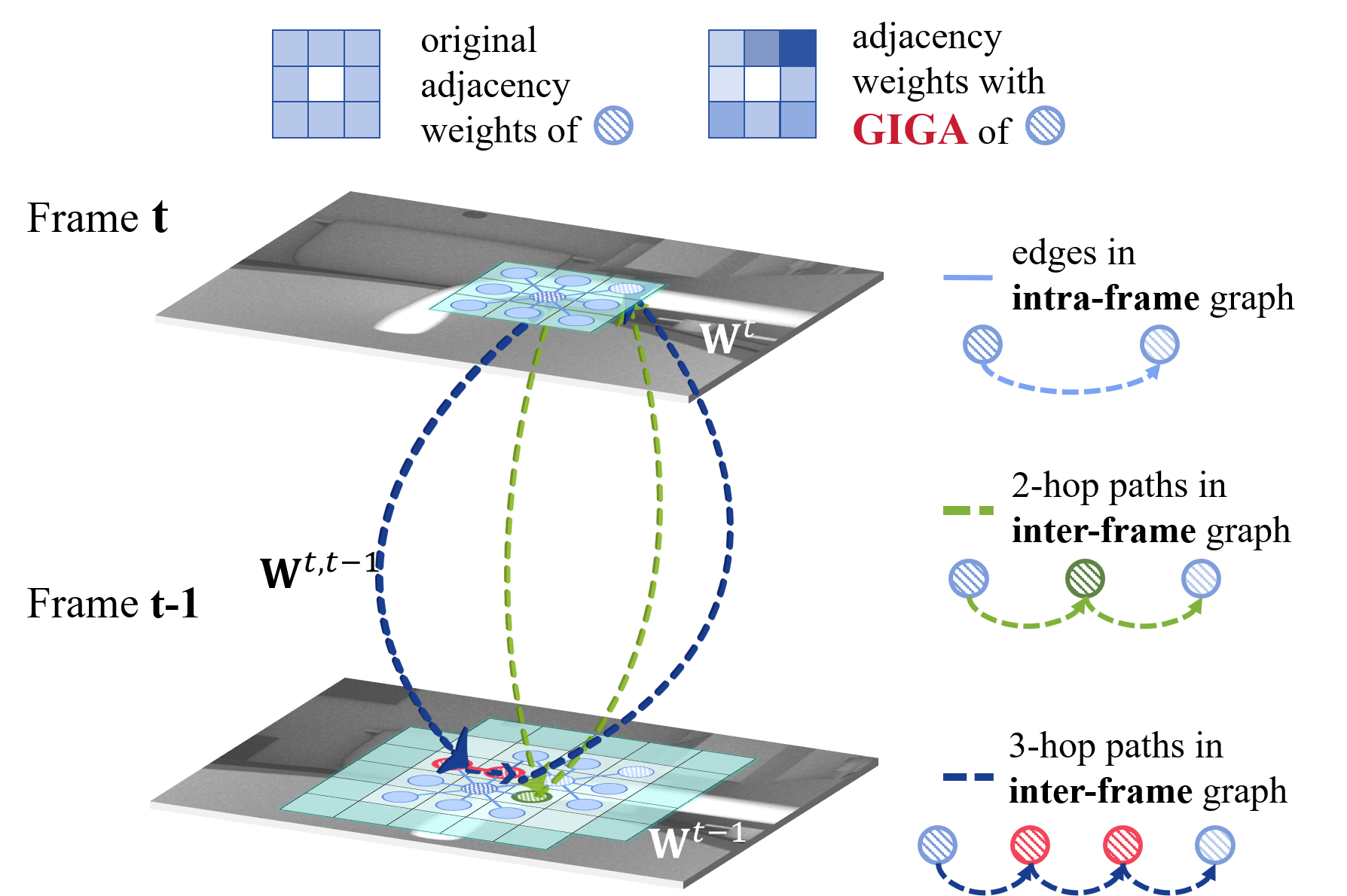}
\vspace{-0.5cm}
\caption{Illustration of cross-frame graph fusion, where the intra-frame graph in reference frame $t-1$ is mapped to current frame $t$ via inter-frame graph with 2-hop or 3-hop paths connecting pixel pairs in frame $t$.
The graph weights are learned in graph-informed geometric attention (GIGA) mechanism, which updates the adjacency matrix in current frame using that in reference frame.}
\label{fig:graph}
\vspace{-0.30cm}
\end{figure}

Since the raw data lie on image grids which naturally defines sparse graph structures \cite{hu2013depth}, we construct 8-connected undirected graphs $\cG^t_i$, $\cG^t_q$ for $\x_i^t$, $\x_q^t$ in each frame, where each pixel is connected to its 8 neighbors as shown in Fig.\,\ref{fig:graph} for frame $t$.
The corresponding non-negative symmetric graph adjacency matrices $\W^t_i, \W^t_q \in \mathbb{R}^{N \times N}$ represent the pair-wise correlation between connected pixels.
For example, 
the $(m,n)$-th element of $\W^t_i$, \ie, $w^t_i(m,n) \geq 0$, indicates the similarity between pixel $m$ and $n$ in $\x_i^t$.

We refer to $\cG^t_i$, $\cG^t_q$ as \textit{intra-frame graphs} which are constructed independently from other frames. 
Nevertheless, the noise corruption in captured ToF raw data may lead to suboptimal graph construction, which motivates us to exploit graph structures in neighboring frames
as auxiliary features to refine graph construction in the current frame.


\subsection{Cross-frame Graph Fusion}
\label{sec:3_inter}

Due to camera motion and object movement/deformation in the dynamic scene, the graph correlations $\W^{t-1}_i$, $\W^{t-1}_q$ in the reference frame need to be mapped to the corresponding pixel pairs before fused with $\W^t_i, \W^t_q$ in the current frame.
To do so, we implement the cross-frame graph mapping as the composition of \textit{intra-frame graph} in frame $t-1$ and \textit{inter-frame graph} between frame $t$ and $t-1$, resulting in the \textit{mapped graph} of frame $t-1$.
Similar to \cite{dong2024exploiting}, we take only the previous frame instead of all the $T$ frames for reference so that multi-frame information is propagated in a forward-only manner.
In the following, we illustrate the mapped graph construction $\hat{\W}^{t-1}_i$ for $\x^t_i$ and hereinafter eliminate the notations $i$ and $q$ since the same procedure applies to the two components.

\noindent \textbf{Inter-frame Graph} For each edge $(m,n)$ in frame $t$, the graph mapping aims to utilize $\W^{t-1}$ for recomputing the edge weight $\hat{w}^t(m,n)$, which is given as the sum of weights of all the possible paths between pixel $m$ and $n$ in the mapped graph.
We first construct inter-frame graph $\W^{t,t-1}$ where each pixel $m$ in frame $t$ is connected to pixels frame $t-1$ within the neighborhood $\cN_m^{t-1}$.
We set $\cN_m^{t-1}$ as a $q \times q$ spatial neighborhood centered at the same coordinate $m$, which is highlighted in green in frame $t-1$ in Fig.\,\ref{fig:graph}.  

\noindent \textbf{Mapped Graph} We construct the mapped graph as the composition of $\W^{t,t-1}$ and $\W^{t-1}$.
To connect $m$ and $n$, there are two types of paths, one is the 2-hop path marked with green dotted lines in Fig.\,\ref{fig:graph},
where $m$ and $n$ are connected via the same pixel $k \in \cN_m^{t-1} \cap \cN_n^{t-1}$, and the corresponding graph weights are computed as $\W^{t,t-1} (\W^{t,t-1})^\top$.
The other type is the 3-hop path marked with blue dotted lines in Fig.\,\ref{fig:graph},
where $m$ and $n$ are connected via the connected pixel pair $(k,l)$ where $k,l \in \cN_m^{t-1} \cap \cN_n^{t-1}$, and the corresponding graph weights are computed as $\W^{t,t-1} \W^{t-1} (\W^{t,t-1})^\top$.

In sum, the mapped graph weights are given as:
\begin{align}\label{eq:w_map}
\hat{\W}^{t-1} & = \W^{t,t-1} (\W^{t,t-1})^\top + \W^{t,t-1} \W^{t-1} (\W^{t,t-1})^\top \nonumber \\
             & = \W^{t,t-1} ( \W^{t-1} + \I ) (\W^{t,t-1})^\top, 
\end{align}
where $\I \in \mathbb{R}^{N \times N}$ is the identity matrix.
Note that $\W^{t,t-1}$ is shared for components $i$ and $q$.

\noindent \textbf{Cross-frame Fused Graph}
Then the cross-frame graph fusion is a weighted average of mapped graph $\hat{\W}^{t-1}$ and original intra-frame graph $\W^t$, resulting in the \textit{fused graph} $\widetilde{\W}^t$:
\begin{align}\label{eq:w_fuse}
\widetilde{\W}^t (\x^t, \x^{t-1}) = \bPhi^{t,t-1} \hat{\W}^{t-1} + \W^t,
\end{align}
where $\widetilde{\W}^t$ depends on frames $\x^t$ and $\x^{t-1}$, and the non-negative diagonal matrix $\bPhi^{t,t-1} \in \mathbb{R}^{N \times N}$ represents the mapping confidence, so as to avoid the effect of inaccurate graph mapping, \eg, in case of occlusion where the mapping is invalid.
Note that the graph edge weights in $\widetilde{\W}^t$ are end-to-end trained as described in Sec.\,\ref{sec:net}.

\subsection{MAP Formulation via Graph Fusion}
\label{sec:3_map}

To denoise $\x_i^t,\x_q^t$ with the captured noisy $\y_i^t,\y_q^t$, we formulate a MAP problem using ToF depth noise distribution to compute likelihood term and image smoothness on fused graph for prior term.

\noindent \textbf{Depth Noise Distribution Induced Likelihood}
First, we compute the distribution of depth noise $\n_d^t$ resulting from noise in $\y_i^t,\y_q^t$. 
As commonly assumed, $\x_i$ and $\x_q$ are corrupted by additive white Gaussian noise (AWGN) \cite{frank2009theoretical,georgiev2018time}, and the pixels in $\y_i^t,\y_q^t$ are independent and identically distributed with multivariate Gaussian distribution.
Based on the depth noise distribution derived in \cite{georgiev2018time,jia2025deep}, we derive the log of likelihood of $\n^t_d$ given as a function of $\x_i^t,\x_q^t$:
\begin{equation} \label{eq:likely}
\ln P(\n^t_d) \approx  -\frac{1}{2\sigma^2} \| (\X_a^t)^{-1}(\x^t_q \odot \y^t_i - \x^t_i \odot \y^t_q) \|_2^2,
\end{equation}
where $\X_a^t = \mathrm{diag}(\x^t_a)$ is the amplitude, $\odot$ is Hadamard product.
Detailed proof of (\ref{eq:likely}) is provided in Sec.\,\ref{sec:likelihood} in the supplementary material.

\noindent \textbf{Graph Smoothness Prior}
Due to the ill-posedness of the problem, extra prior knowledge describing the characteristics of $\x_i^t,\x_q^t$ is required to facilitate the reconstruction.
Here, we adopt the widely used graph Laplacian regularization (GLR) prior \cite{pang2017graph} to impose image smoothness on the cross-frame fused graph given as:
\begin{align} \nonumber
P(\x^t_i, \x^t_q) = \exp &( - \frac{(\x_i^t)^\top \widetilde{\L}_i^t (\x_i^t, \x_i^{t-1}) \x_i^t }{\sigma_L^2} ) \\ \label{eq:GLR}
&\times \exp ( - \frac{ (\x^t_q)^\top \widetilde{\L}_q^t (\x_q^t, \x_q^{t-1}) \x^t_q}{\sigma_L^2} ),
\end{align}
where $\sigma_L$ adjusts the sensitivity to variations on graphs, and the fused graph Laplacian matrix $\widetilde{\L}_i^t(\x_i^t, \x_i^{t-1})$ is given as:
\begin{align} 
    &\widetilde{\L}_i^t(\x_i^t, \x_i^{t-1}) = \widetilde{\D}_i^t (\x_i^t, \x_i^{t-1}) - \widetilde{\W}_i^t (\x_i^t, \x_i^{t-1}), \\
    &\widetilde{\D}_i^t (\x_i^t, \x_i^{t-1}) = \mathrm{diag}(\widetilde{\W}_i^t (\x_i^t, \x_i^{t-1}) \mathbf{1}), \label{eq:deg}
\end{align}
where $\mathbf{1} \in \mathbb{R}^N$ is an all-one vector. $\widetilde{\L}_q^t (\x_q^t, \x_q^{t-1})$ is computed with the same procedure.

\begin{figure*}[t]
\centering
\includegraphics[width=\textwidth]{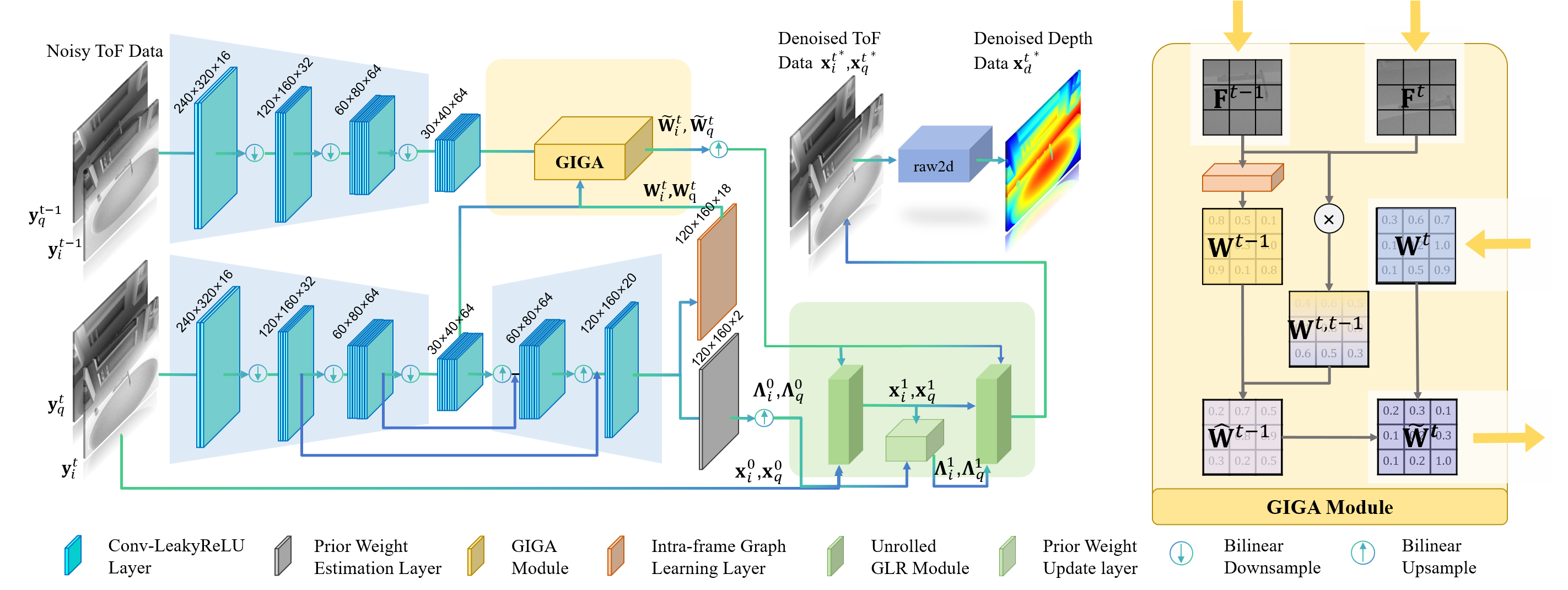}
\vspace{-0.5cm}
\caption{Framework of GIGA-ToF network which is composed of (1) the feature extraction network in blue to extract geometric features from ToF raw data and estimate initial prior weights and intra-graph adjacency matrices, (2) Graph-Induced Geometric Attention (GIGA) module in yellow to learn graph edges from the geometric features informed by graph structures as shown at the right side, and (3) Unrolled GLR module in green to denoise ToF data. Output dimensions are shown on top of each layer.}
\label{fig:framework}
\vspace{-0.30cm}
\end{figure*}

\noindent \textbf{MAP Formulation}
The MAP problem is formulated based on (\ref{eq:likely}) and (\ref{eq:GLR}) and is given as:
\begin{align} \nonumber
&\min_{\x^t_i, \x^t_q} \quad   
\frac{1}{2\sigma^2} \| (\X_a^t)^{-1}(\x^t_q \odot \y^t_i - \x^t_i \odot \y^t_q) \|_2^2 \\ \label{eq:obj}
&+ \frac{(\x_i^t)^\top \widetilde{\L}_i^t (\x_i^t, \x_i^{t-1}) \x_i^t }{\sigma_L^2} + \frac{(\x^t_q)^\top \widetilde{\L}_q^t (\x_q^t, \x_q^{t-1}) \x^t_q}{\sigma_L^2},
\end{align}
so that ToF raw data in frame $t$ is denoised towards \textit{temporal consistency} between frame $t$ and $t-1$ \textit{by utilizing cross-frame graph fusion}.
(\ref{eq:obj}) is then approximately solved with alternating optimization.
In each iteration, we fix $\x^t_q$ and optimize $\x^t_i$, then fix $\x^t_i$ and optimize $\x^t_q$, and repeat until convergence.
For example, in iteration $r$, we set $\x_a^t = \x_a^{t,r-1}$, then fix $\x^t_q = \y^t_q = \x^{t,r-1}_q$ and optimize $\x^t_i$ as
\begin{align} \nonumber
\min_{\x^t_i} \quad & ||  (\X_a^{t,r-1})^{-1}\x_q^{t,r-1} \odot (\x^t_i - \x_i^{t,r-1}) ||_2^2 \\ \label{eq:obj_i}
& + 2\lambda (\x^t_i)^\top \widetilde{\L}_i^t (\x_i^t, \x_i^{t-1}) \x^t_i,
\end{align}
where $\lambda=(\sigma/\sigma_L)^2$.
$\x_i^{t,0},\x_q^{t,0}$ are initialized with $\y_i^t,\y_q^t$ in the first iteration.
The remaining questions are 1) how to efficiently solve (\ref{eq:obj_i}) and 2) how to learn fused graph from data, which are addressed as follows.

\section{Network Architecture}
\label{sec:net}

The graph-based solution to (\ref{eq:obj_i}) is unrolled into iterative convolutional filtering with kernels learned from graph-informed geometric attention in Sec.\,\ref{sec:4_giga}, which induces the interpretable network design in Sec.\,\ref{sec:4_net}, enhancing network robustness to cross-dataset generalization.

\subsection{Algorithm Unrolling and Graph Learning}
\label{sec:4_giga}

By differentiating (\ref{eq:obj_i}) with respect to $\x^t_i$ and setting the result equal to 0, we get the solution by solving the following linear system,
\begin{align} \nonumber
((\X_a^{t,r-1})^{-1} | \x_q^{t,r-1} |)^2 &\odot (\x^t_i - \x_i^{t,r-1}) \\ 
&+ 2\lambda \widetilde{\L}_i^t (\x_i^t, \x_i^{t-1}) \x^t_i = 0. \label{eq:qp_system0}
\end{align}

\noindent \textbf{Unrolled GLR}
For accurate estimation of parameters $\lambda$ and $\widetilde{\L}_i^t (\x_i^t, \x_i^{t-1})$, we follow \cite{jia2025deep} and unroll the solution of (\ref{eq:qp_system0}) into iterative filtering based on gradient descent, so that the parameters are fully trainable with DNN.
Specifically, starting with $\x_i^{t,r,0} = \x_i^{t,r-1}$, the solution is given by running the following solution procedure,
\begin{align}
    &\x_i^{t,r,p+1} = \frac{\x_i^{t,r-1}  + \bLambda_i^{t,r-1} \widetilde{\W}_i^t (\x^t, \x^{t-1}) \x_i^{t,r,p}}{\I + \bLambda_i^{t,r-1} \widetilde{\D}_i^t (\x^t, \x^{t-1})},  \label{eq:local_op} \\
    & \bLambda_i^{t,r-1} = 2\lambda \left(\text{diag}(|\x_q^{t,r-1}|)^{-1}\X_a^{t,r-1}\right)^2, \label{eq:local_a}
\end{align}
where $\bLambda_i^{t,r-1}$ is a diagonal matrix and can be considered as the pixel-wise weighting factor for the GLR prior. 
In (\ref{eq:local_op}), the $(p+1)$-th iteration output is computed via a convolutional transform of $p$-th iteration result $\x_i^{t,r,p}$ with kernel $\widetilde{\W}_i^t (\x^t, \x^{t-1})$, followed by a fusion with input $\x_i^{t,r-1}$ with weight $\bLambda_i^{t,r-1}$.
By recurrently repeating the above procedure, we obtain the solution to (\ref{eq:local_op}), which is summarized in Algorithm \ref{algo:giga} in Sec.\,\ref{sec:al_sum} in the supplementary material.

\noindent \textbf{Graph-Informed Geometric Attention}
Next, we discuss graph learning to compute edge weights in $\widetilde{\W}_i^t$ and $\widetilde{\W}_q^t$. 
In the following, we illustrate the estimation of $\widetilde{\W}_i^t$ for $\x^t_i$ and hereinafter eliminate the notations $i$ and $q$ since the same procedure applies to the two components.

First, for intra-frame graph learning, we use the geometric features from ToF raw data in frames $t$ and $t-1$, \ie, $\F^t$ and $\F^{t-1}$ to estimate $\W^t$ and $\W^{t-1}$ with a single convolution layer.
Then, to compute inter-frame graph $\W^{t,t-1}$, we adopt a variant of the basic self-attention operation for graph computation following \cite{do2025interpretable}, where attention weight is computed as,
\begin{equation}
    a_{ij}= \mathrm{softmax}( e_{ij} ),\quad e_{ij} = (\Q\F^t(i))^{\top} (\K\F^{t-1}(j)),
\end{equation}
where $\Q,\K \in \mathbb{R}^{C \times C}$ are the query and key
matrices, respectively, and $C$ is the feature dimension.
Then the mapped and fused graphs are computed using (\ref{eq:w_map}) and (\ref{eq:w_fuse}).
The above procedure for graph learning is named Graph-Induced Geometric Attention (GIGA) module to learn graph edges from the geometric features informed by graph structures as illustrated at the right side of Fig.\,\ref{fig:framework}.


\subsection{GIGA-ToF Network Architecture}
\label{sec:4_net}

Leveraging the GIGA module in Sec.\,\ref{sec:4_giga}, we propose the \textit{GIGA-ToF} network for ToF depth denoising which is composed of three parts as shown in Fig.\,\ref{fig:framework}.
The first part is the feature extraction network that adopts an encoder-decoder structure with skip-connections \cite{ronneberger2015u} to estimate multi-scale features of scales $s \in \{1/8, 1/4, 1/2 \}$, where the feature dimensions are shown in Fig.\,\ref{fig:framework}.
$\F^t$ at scale $1/2$ is used to estimate initial prior weights $\bLambda_i^{t,0},\bLambda_q^{t,0}$ via a 1-layer convolution.
We apply sigmoid function on $\bLambda_i^{t,0},\bLambda_q^{t,0}$ to get positive weights, then scale by $10$ to ensure sufficient denoising strength.
All the convolutional layers adopt $3\times3$ kernel size with LeakyReLU activation.

The second part is the GIGA module. The features $\F^{t-1},\F^t$ at scale $1/8$ are fed into the GIGA module to compute $\W^{t-1}$, $\W^{t,t-1}$ and $\bPhi^{t,t-1}$ for computational efficiency, which generates $\hat{\W}^{t-1}$ at $1/8$ scale.
The neighborhood size for inter-frame graph is set as $q=7$.
For detail refinement, $\F^t$ at scale $1/2$ is used for computing $\W^t$, which is fused with bilinear upsampled $\hat{\W}^{t-1}$.

The third part is the unrolled GLR module adopted from \cite{jia2025deep}.
The final output $\x_i^{t,*}$ and $\x_q^{t,*}$ are converted to depth $\x_d^{t,*}$ via the raw2d module based on (\ref{eq:d}).
In the case of multi-frequency inputs, raw data of different $f_m$ are denoised separately with shared network parameters.
Depth maps with different $f_m$ are merged via phase unwrapping \cite{su2018deep} to generate the final depth. 

\noindent \textbf{Graph Spectral Filtering Interpretability}
Since the graph Laplacian matrix in (\ref{eq:obj_i}) is symmetric and positive semi-definite (PSD) with positive edge weight, its solution is a low-pass graph spectral filtering. 
Therefore, together with the graph spectral filtering interpretability and the incorporation of ToF imaging mechanism in the network design, the proposed GIGA-ToF is fully interpretable which effectively enhances its robustness to cross-dataset generalization as validated in Sec.\,\ref{sec:kinect}.

\subsection{Loss Function}
We train our network with $l_1$ loss function supervised by the ground truth $\x_i^{t,\text{gt}}$, $\x_q^{t,\text{gt}}$ as follows:
\begin{equation} \label{eq:loss}
L = \frac{1}{|\cV|} \sum_{v \in \cV}  \sum_{\theta \in \{i,q\}} \left|\x_{\theta}^{t,*}(v) - \x_{\theta}^{t,\text{gt}}(v) \right| , 
\end{equation} 
where $v$, $\cV$ and $|\cV|$ denote the pixel index, set of valid pixels in GT, and the number of valid pixels, respectively. 
\begin{table*}[t]
\centering
\caption{Comparison of denoising accuracy on synthetic DVToF testing dataset and augmented dataset}
\vspace{-0.3cm}
\resizebox{2.0\columnwidth}{!}{ 
\begin{tabular}{lcccccccccc}
\toprule
Methods & Runtime & Memory & \multicolumn{4}{c}{DVToF Dataset} & \multicolumn{4}{c}{DVToF Dataset with augmented noise} \\
\cmidrule(lr){2-3} \cmidrule(lr){4-7} \cmidrule(lr){8-11}
&  (s)    &   (MB) & MAE(m)$\downarrow$ & AbsRel$\downarrow$ & $\delta_1$$\uparrow$ & TEPE(m)$\downarrow$ & MAE(m)$\downarrow$ & AbsRel$\downarrow$ & $\delta_1$$\uparrow$ & TEPE(m)$\downarrow$\\
\midrule
\textit{\textbf{Single-frame}}\\
libfreenect2 \cite{xiang2016libfreenect2} & 0.003 & - & 0.1044 & 0.0283 & 0.9746 & 0.1023 & 0.1230 & 0.0386 & 0.9645 & 0.1234 \\
DeepToF \cite{marco2017deeptof} & 0.006 & 738 & 0.2172 & 0.1071 & 0.8951 & 0.2003 & 0.2830 & 0.1409 & 0.8534 & 0.2705 \\
ToFNet \cite{su2018deep} & 0.008 & 1468 & 0.1290 & 0.0652 & 0.9586 & 0.1221 & 0.1334 & 0.0677 & 0.9564 & 0.1275 \\
UDA \cite{agresti2022unsupervised} & 0.006 & 900 & 0.0564 & 0.0152 & 0.9880 & 0.0884 & 0.1153 & 0.0570 & 0.9451 & 0.1274 \\
RADU \cite{schelling2022radu} & 83.7 & 11115 &  0.1350 & 0.0697 & 0.9497 &  0.1290 & 0.1264 & 0.0610 & 0.9623 & 0.1202 \\
GLRUN \cite{jia2025deep} & 0.016 & 766 & 0.0357 & 0.0107 & 0.9929 & 0.0734 & 0.0550 & 0.0244 & 0.9896 & 0.1221\\
\midrule
\textit{\textbf{Multi-frame}} \\
WMF \cite{min2011depth}   &   24.3 & - &0.0311 & 0.0116 & 0.9955 &0.0751 &0.0495 &0.0209 &0.9898& \textbf{0.0950}\\
ConvLSTM \cite{patil2020don} & 0.019 & 1362 & 0.1314 &0.0337 &0.9624& 0.1143 &  0.1257 &0.0406 &0.9736& 0.1411\\
DVSR \cite{sun2023consistent} & 0.632 & 1308 &0.0718 &0.0844& 0.9777 &0.1176& 0.0791 &0.0425& 0.9736 &0.1271\\
MTDNet \cite{dong2024exploiting}  & 0.584 & 317  &0.0566 &0.0642& 0.9816&0.1046 &0.0625 &0.0316 &0.9778 &0.1129\\
\ourmethod (Ours) & 0.027 & 824 & \textbf{0.0193} & \textbf{0.0060} & \textbf{0.9974} & \textbf{0.0637} & \textbf{0.0487} & \textbf{0.0205} &\textbf{0.9903} & 0.1102 \\
\bottomrule
\end{tabular}
}
\label{tab:comparison}
\end{table*}

\begin{figure*}[t]
\centering
\includegraphics[width=\textwidth]{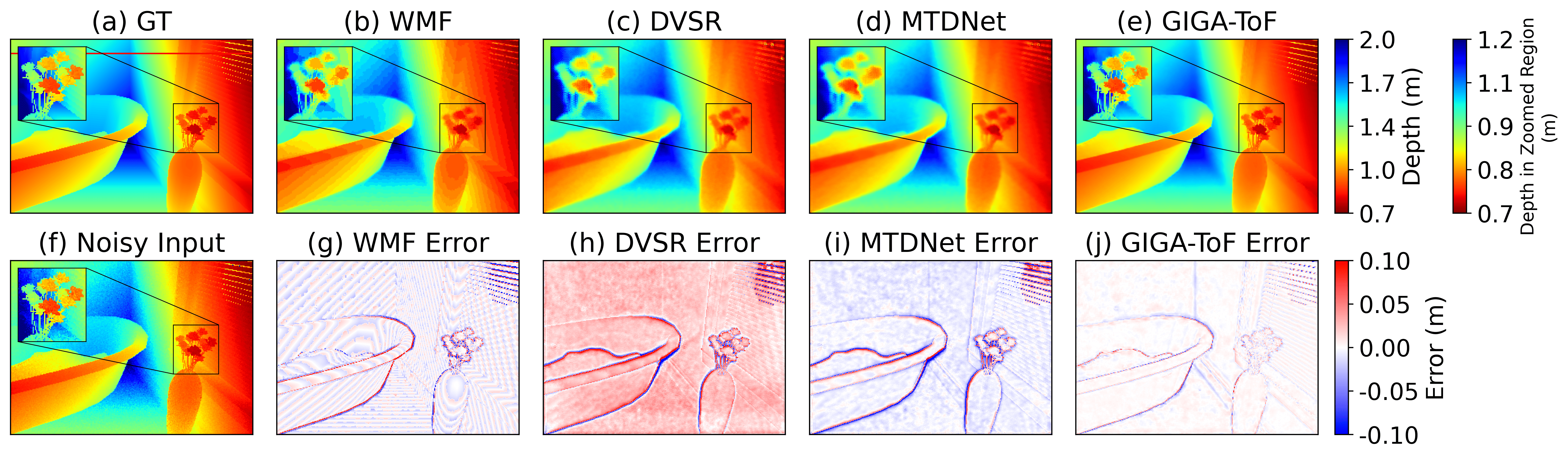}
\vspace{-0.6cm}
\caption{Depth results and error maps of ToF depth denoised on DvToF dataset: (a) GT, results of (b) WMF \cite{min2011depth}, (c) DVSR \cite{sun2023consistent}, (d) MTDNet \cite{dong2024exploiting} and (e) proposed GIGA-ToF. Corresponding error maps are in the second row.}
\label{fig:exp_dvtof}
\vspace{-0.3cm}
\end{figure*}

\section{Experimental Results}
\label{sec:exp}
We first generate syntheic DVToF data with temporal ToF data and depth, which is used for training.
Then we evaluate the network performance with DVToF testing data, and further show generalization to real Kinectv2 depth images.

\subsection{Experimental Settings}
\noindent \textbf{Datasets} We adopted the dataset generation protocol in \cite{su2018deep} while the camera paths were randomly generated to augment the cross-frame flows. We have 5 static scenes, each with 10 paths of 250-frame length, generating 12.5$k$ measurements of raw ToF correlation-depth pairs in total with resolution $320\times240$. 
The resulting dataset is named \textit{DVToF} which stands for \textit{depth video of ToF data}.
In addition, we generated random noise using Kinectv2 noise statistics provided in \cite{guo2018tackling}.
We used 9375 pairs for training and 3125 pairs with unseen scene-path configurations for testing. 
More importantly, to evaluate with real data, we captured real ToF data with Kinectv2 camera and applied the pre-trained model on DVToF dataset to evaluate the cross-dataset generalization ability.

\noindent \textbf{Training details} We used Adam optimizer with initial learning rate $1e^{-3}$ and decay at epoch $[15, 30, 45]$ with decay rate $0.7$. 
The model was trained from scratch for 60 epochs.
We employed the PyTorch framework \cite{paszke2017automatic} 
on a single GeForce RTX 3090 GPU.
We set $T=3$ and $R=2$ for the Unrolled GLR module.

\noindent \textbf{Metrics} 
Following \cite{sun2023consistent}, we used per-frame mean absolute error (MAE), Absolute Relative Error (AbsRel), and $\delta_1$ accuracy to evaluate per-frame depth estimation accuracy; and temporal end-point error (TEPE) to measure temporal consistency.
For complexity comparison, we tested the average runtime and GPU memory cost with DVToF dataset using single 3090 GPU and Intel i9-14900K CPU. 
We did not report memory costs for methods running only on CPU.

\subsection{Comparison with Existing Schemes}
We compared with the following competing schemes. 
\begin{itemize}
    \item Model-based methods: single-frame libfreenect2 \cite{xiang2016libfreenect2} and multi-frame weighted mode filter (WMF) \cite{min2011depth};
    \item Single-frame based DNNs: DeepToF \cite{marco2017deeptof}, ToFNet \cite{su2018deep}, UDA \cite{agresti2019unsupervised}, RADU \cite{schelling2022radu}, GLRUN \cite{jia2025deep};
    \item Multi-frame DNNs: ConvLSTM \cite{patil2020don}, DVSR \cite{sun2023consistent}, MTDNet \cite{dong2024exploiting}. 
\end{itemize}
To ensure a fair comparison, all competing methods were retrained and tested on the DVToF dataset. In addition, following \cite{barron2016fast}, we augmented the DVToF dataset with simulated edge noise. Note that the same model was used for testing in both noise settings to test generalization ability to unseen noise. 
As shown in Table.~\ref{tab:comparison}, GIGA-ToF achieves the best accuracy performance in both noise settings, outperforming other methods by at least \textbf{37.9\%} in MAE and \textbf{13.2\%} in TEPE in normal noise setting.
Also, the complexity of GIGA-ToF is moderate among SOTA methods, while the competing WMF is computationally costly and hinders its application in real-time usage.

For visual evaluation, we present qualitative comparison of multi-frame methods in Fig.~\ref{fig:exp_dvtof}. GIGA-ToF generates smooth results while preserving fine details, as highlighted in the zoomed-in region, further confirming GIGA-ToF’s ability to maintain spatial sharpness while effectively removing noise due to motion-invariant graph structure fusion.
Note that WMF generates better TEPE in the barron noise setting, showing competing temporal consistency, but suffers from quantization error as shown in Fig.~\ref{fig:exp_dvtof}(b).
In addition, DNN-based DVSR and MTDNet show blurry details due to the fusion of temporally varying depth features.

\begin{figure}[t]
\centering
\includegraphics[width=0.9\columnwidth]{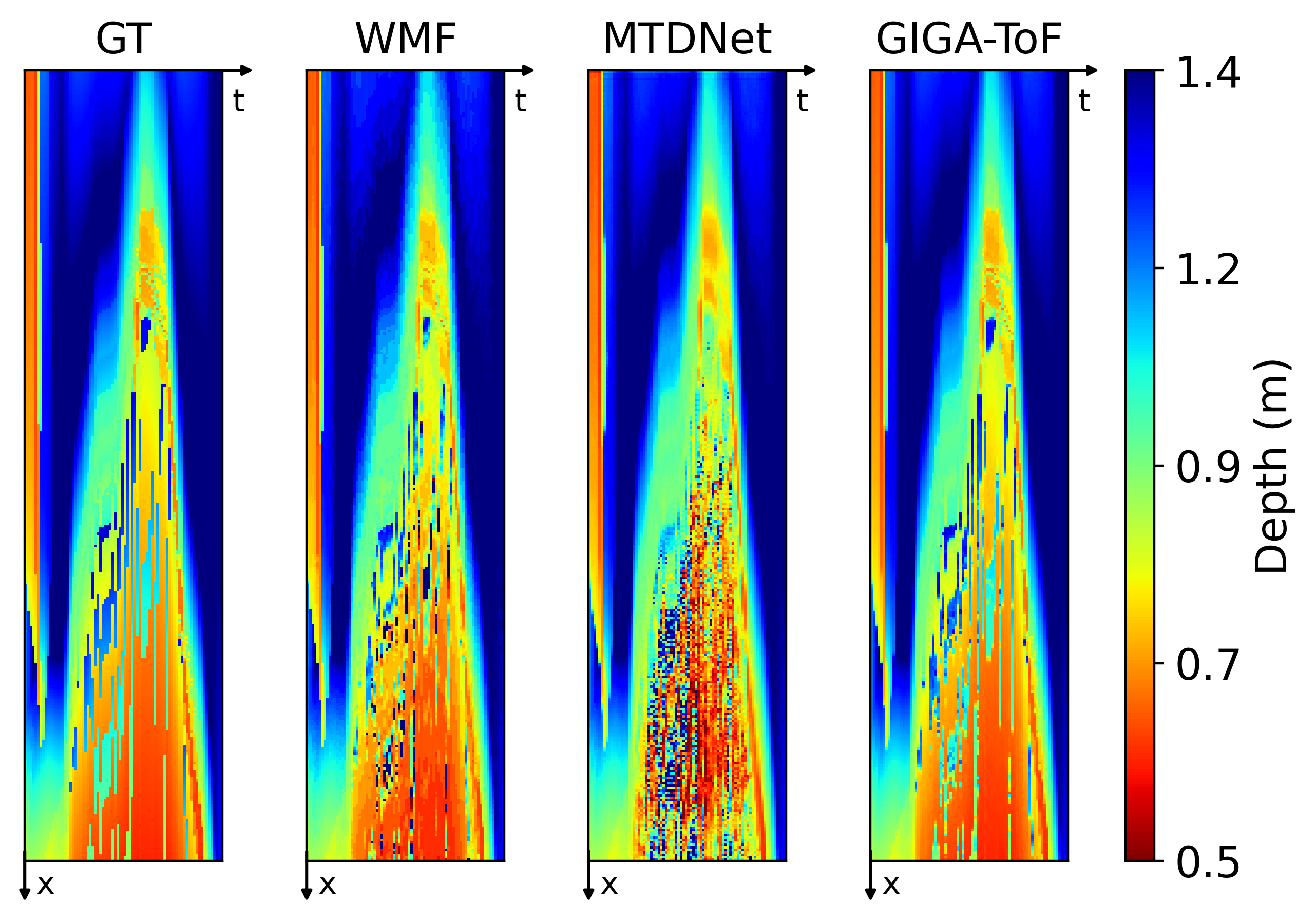}
\vspace{-0.4cm}
\caption{x-t slices (along red line in Fig.~\ref{fig:exp_dvtof}(a)) for temporal stability visualization, where GIGA-ToF exhibits clear details and less noise than competing multi-frame schemes.}
\label{fig:x_t}
\vspace{-0.30cm}
\end{figure}

To visualize temporal consistency, we plot the x-t slices of the estimated depth images of multi-frame methods in Fig.~\ref{fig:x_t}.
While MTDNet and WMF remain noisy with noticeable temporal jittering, GIGA-ToF exhibits clean x-t slices, demonstrating its high temporal consistency.
Please refer to the supplementary video for better temporal visualizations.

\begin{figure*}[t]
\centering
\includegraphics[width=\textwidth]{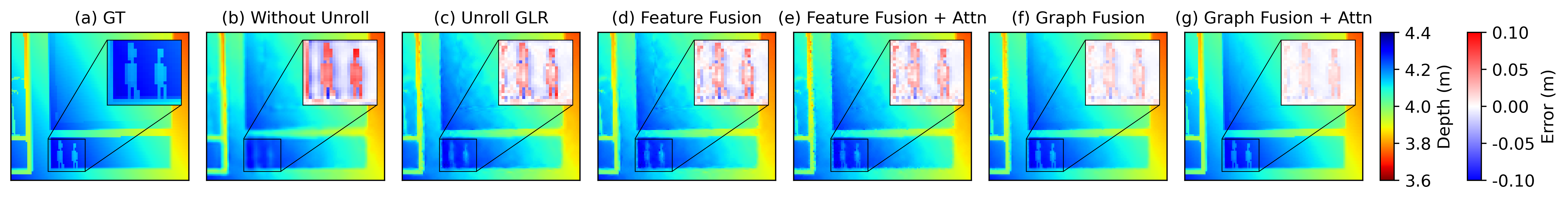}
\vspace{-0.6cm}
\caption{Comparison of denoising results under different GIGA-ToF variants. (a) GT, results (b) without and (c) with unrolled GLR in single-frame processing, feature fusion (d) without and (e) with attention; graph structure fusion (f) without and (g) with attention.}
\label{fig:exp_ab}
\vspace{-0.30cm}
\end{figure*}

\subsection{Generalization to Real Data}
\label{sec:kinect}

\begin{figure}[t]
\centering
\includegraphics[width=\columnwidth]{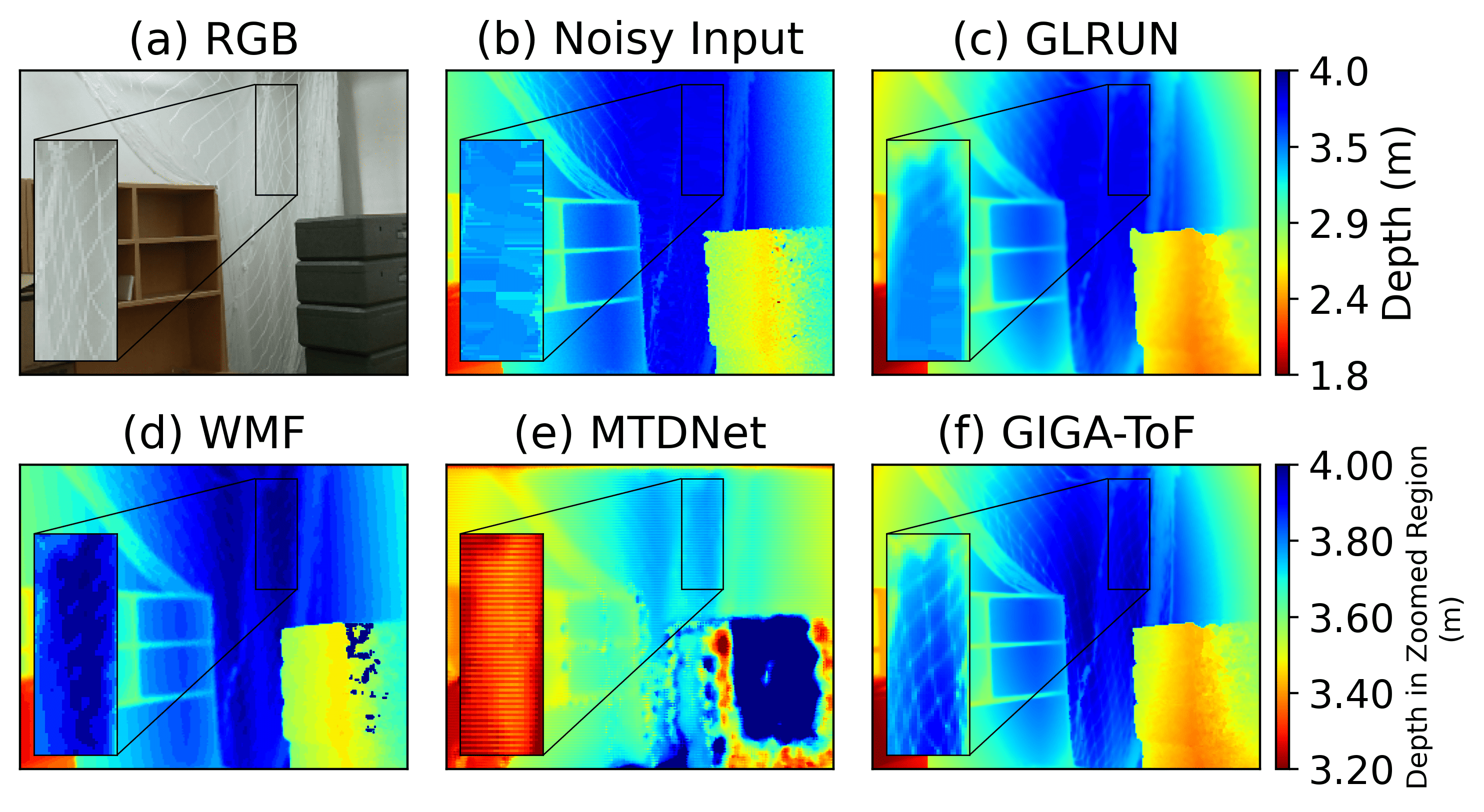}
\vspace{-0.6cm}
\caption{Visual results of ToF depth denoising on real data captured by Kinect v2 sensor: (a) RGB and (b) noisy depth captured by Kinectv2 camera, and results of (c) GLRUN, (d) WMF, (e) MTDNet and (f) GIGA-ToF, where GIGA-ToF shows robustness to real noise and recovers accurate details.}
\label{fig:kinect}
\vspace{-0.3cm}
\end{figure}

To assess generalization ability of GIGA-ToF on real-world data, we capture ToF data with Kinect v2 camera \cite{kurillo2022evaluating} and conduct qualitative comparison shown in Fig.~\ref{fig:kinect}. 
While DNN-based MTDNet fails to generalize to real data, model-based WMF shows stable but blurry results.
While GLRUN shows robustness to real data, multi-frame processing GIGA-ToF further enhances the detail preservation, which validates the necessity of utilizing temporal correlation.
In sum, GIGA-ToF, despite being trained on synthetic data, shows strong generalization to real-world data. 

\subsection{Ablation Study}

To investigate the effectiveness of each component in GIGA-ToF, we test on DVToF dataset with different variants of GIGA-ToF. 
Quantitative results in Table~\,\ref{tab:exp_ablation} and qualitative results in Fig.\,\ref{fig:exp_ab} validate the effectiveness of each component for denoising accuracy and stability.

\noindent\textbf{Unrolled GLR}
In single-frame processing settings, we compare variants with and without Unrolled GLR module. 
First, single-frame variants are much more noisy than multi-frame variants, validating the necessity of temporal processing.
In addition, by removing Unrolled GLR module, the results become much more blurry, validating the effect of graph structure in detail preservation.  

\noindent\textbf{Fusion Mechanism}
For multi-frame processing setting, we investigate the two fusion mechanisms, \ie, depth features and graph structures.
For depth feature fusion, the current and reference frame features at scale $1/8$ are fused in the feature extraction network, where the graph construction is based on the fused features. 
Graph-based fusion outperforms those based on depth feature fusion and exhibits sharper details, validating the effect of motion-invariant graph fusion in resolving spatial ambiguity.

\noindent\textbf{Inter-frame Attention}
In addition, we investigate the effect of inter-frame attention in fusing cross-frame features.
Without using attention for fusion, the features in reference are fused into current frame indifferently, resulting in noticeable noise in the results due to inaccurate fusion correspondence between frames.
This validates the effect of attention in mapping geometric features with accurate correspondence.

\begin{table}[t]
\centering
\fontsize{8.0}{10}\selectfont
\caption{Comparison of quantitative evaluation on DvToF testing dataset with GIGA-ToF variants}
\vspace{-0.1cm}
\label{tab:exp_ablation}
\begin{tabular}{ccc|cccc}
\hline
\multicolumn{3}{c|}{Modules} & MAE & AbsRel & $\delta_1$& TEPE  \\
\cline{1-3}
GLR & Fusion & Attn & (m)$\downarrow$ & $\downarrow$  &  $\uparrow$& (m)$\downarrow$ \\ 
\hline
- & - & - &  0.0409 & 0.0174 & 0.9909 & 0.0793\\
Unroll & - & - &  0.0357 & 0.0107 & 0.9929 & 0.0734 \\
\hline
Unroll &Feature & - & 0.0238 &0.0078& 0.9965 &0.0718 \\
Unroll &Feature & \checkmark & 0.0214  &0.0069 &0.9969 &0.0713 \\
\hline
Unroll &Graph & - & 0.0219 & 0.0078& 0.9970  &0.0702\\
Unroll &Graph & \checkmark &\textbf{0.0193} & \textbf{0.0060} & \textbf{0.9974} & \textbf{0.0637}\\
\hline
\end{tabular}
\vspace{-0.1cm}
\end{table}

\subsection{Limitation and Future Work}
In the current setting, we only consider the previous frame for reference, while the features in more previous frames are not fully utilized. Although involving two frames for multi-frame processing already boosts the depth accuracy and produces temporally consistent results, extending to more frames has not yet been explored.
Therefore, for future study, the investigation will be devoted to a more general processing pipeline for varying input sequence length with recurrent network design.


\section{Conclusion}
\label{sec:con}

In this paper, we propose GIGA-ToF network for ToF depth denoising, simultaneously enhancing temporal consistency and spatial sharpness  utilizing the motion-invariant graph structures.
Based on the cross-frame graph fusion, we impose image smoothness as a prior in the MAP formulation, which is efficiently optimized via algorithm unrolling to produce high-performance yet interpretable network designs.
The resulting network shows enhanced denoising accuracy on synthetic DVToF dataset and higher robustness to real noise over competing schemes due to the graph spectral filter interpretation. 

{
    \small
    \bibliographystyle{ieeenat_fullname}
    \bibliography{refs}

\begin{thebibliography}{41}
\providecommand{\natexlab}[1]{#1}
\providecommand{\url}[1]{\texttt{#1}}
\expandafter\ifx\csname urlstyle\endcsname\relax
  \providecommand{\doi}[1]{doi: #1}\else
  \providecommand{\doi}{doi: \begingroup \urlstyle{rm}\Url}\fi

\bibitem[Agresti et~al.(2019)Agresti, Schaefer, Sartor, and Zanuttigh]{agresti2019unsupervised}
Gianluca Agresti, Henrik Schaefer, Piergiorgio Sartor, and Pietro Zanuttigh.
\newblock Unsupervised domain adaptation for tof data denoising with adversarial learning.
\newblock In \emph{Proceedings of the IEEE/CVF Conference on Computer Vision and Pattern Recognition}, pages 5584--5593, 2019.

\bibitem[Agresti et~al.(2022)Agresti, Schäfer, Sartor, Incesu, and Zanuttigh]{agresti2022unsupervised}
Gianluca Agresti, Henrik Schäfer, Piergiorgio Sartor, Yalcin Incesu, and Pietro Zanuttigh.
\newblock Unsupervised domain adaptation of deep networks for tof depth refinement.
\newblock \emph{IEEE Transactions on Pattern Analysis and Machine Intelligence}, 44\penalty0 (12):\penalty0 9195--9208, 2022.

\bibitem[Barron and Poole(2016)]{barron2016fast}
Jonathan~T Barron and Ben Poole.
\newblock The fast bilateral solver.
\newblock In \emph{European conference on computer vision}, pages 617--632. Springer, 2016.

\bibitem[Bhandari and Raskar(2016)]{bhandari2016signal}
Ayush Bhandari and Ramesh Raskar.
\newblock Signal processing for time-of-flight imaging sensors: An introduction to inverse problems in computational 3-d imaging.
\newblock \emph{IEEE Signal Processing Magazine}, 33\penalty0 (5):\penalty0 45--58, 2016.

\bibitem[Chen et~al.(2022)Chen, Ying, Xue, Wen, and Liu]{chen2022configurable}
Faquan Chen, Rendong Ying, Jianwei Xue, Fei Wen, and Peilin Liu.
\newblock A configurable and real-time multi-frequency 3d image signal processor for indirect time-of-flight sensors.
\newblock \emph{IEEE Sensors Journal}, 22\penalty0 (8):\penalty0 7834--7845, 2022.

\bibitem[Chen et~al.(2020)Chen, Ren, Cheng, Qian, Wang, and Gu]{chen2020very}
Yan Chen, Jimmy Ren, Xuanye Cheng, Keyuan Qian, Luyang Wang, and Jinwei Gu.
\newblock Very power efficient neural time-of-flight.
\newblock In \emph{Proceedings of the IEEE/CVF Winter Conference on Applications of Computer Vision}, pages 2257--2266, 2020.

\bibitem[Cheung et~al.(2018)Cheung, Magli, Tanaka, and Ng]{cheung2018graph}
Gene Cheung, Enrico Magli, Yuichi Tanaka, and Michael~K Ng.
\newblock Graph spectral image processing.
\newblock \emph{Proceedings of the IEEE}, 106\penalty0 (5):\penalty0 907--930, 2018.

\bibitem[Do et~al.(2025)Do, Eftekhar, Hosseini, Cheung, and Chou]{do2025interpretable}
Tam~Thuc Do, Parham Eftekhar, Seyed~Alireza Hosseini, Gene Cheung, and Philip Chou.
\newblock Interpretable lightweight transformer via unrolling of learned graph smoothness priors.
\newblock \emph{Advances in Neural Information Processing Systems}, 37:\penalty0 6393--6416, 2025.

\bibitem[Dong et~al.(2024)Dong, Zhang, Sun, and Xiong]{dong2024exploiting}
Guanting Dong, Yueyi Zhang, Xiaoyan Sun, and Zhiwei Xiong.
\newblock Exploiting dual-correlation for multi-frame time-of-flight denoising.
\newblock In \emph{European Conference on Computer Vision}, pages 473--489. Springer, 2024.

\bibitem[Du et~al.(2020)Du, Turner, Dzitsiuk, Prasso, Duarte, Dourgarian, Afonso, Pascoal, Gladstone, Cruces, et~al.]{du2020depthlab}
Ruofei Du, Eric Turner, Maksym Dzitsiuk, Luca Prasso, Ivo Duarte, Jason Dourgarian, Joao Afonso, Jose Pascoal, Josh Gladstone, Nuno Cruces, et~al.
\newblock Depthlab: Real-time 3d interaction with depth maps for mobile augmented reality.
\newblock In \emph{Proceedings of the 33rd Annual ACM Symposium on User Interface Software and Technology}, pages 829--843, 2020.

\bibitem[Frank et~al.(2009{\natexlab{a}})Frank, Plaue, and Hamprecht]{frank2009denoising}
Mario Frank, Matthias Plaue, and Fred~A Hamprecht.
\newblock Denoising of continuous-wave time-of-flight depth images using confidence measures.
\newblock \emph{Optical Engineering}, 48\penalty0 (7):\penalty0 077003--077003, 2009{\natexlab{a}}.

\bibitem[Frank et~al.(2009{\natexlab{b}})Frank, Plaue, Rapp, K{\"o}the, J{\"a}hne, and Hamprecht]{frank2009theoretical}
Mario Frank, Matthias Plaue, Holger Rapp, Ullrich K{\"o}the, Bernd J{\"a}hne, and Fred~A Hamprecht.
\newblock Theoretical and experimental error analysis of continuous-wave time-of-flight range cameras.
\newblock \emph{Optical Engineering}, 48\penalty0 (1):\penalty0 013602--013602, 2009{\natexlab{b}}.

\bibitem[Georgiev et~al.(2018)Georgiev, Bregovi{\'c}, and Gotchev]{georgiev2018time}
Mihail Georgiev, Robert Bregovi{\'c}, and Atanas Gotchev.
\newblock Time-of-flight range measurement in low-sensing environment: Noise analysis and complex-domain non-local denoising.
\newblock \emph{IEEE Transactions on Image Processing}, 27\penalty0 (6):\penalty0 2911--2926, 2018.

\bibitem[Guo et~al.(2018)Guo, Frosio, Gallo, Zickler, and Kautz]{guo2018tackling}
Qi Guo, Iuri Frosio, Orazio Gallo, Todd Zickler, and Jan Kautz.
\newblock Tackling 3d tof artifacts through learning and the flat dataset.
\newblock In \emph{Proceedings of the European Conference on Computer Vision (ECCV)}, pages 368--383, 2018.

\bibitem[Gutierrez-Barragan et~al.(2021)Gutierrez-Barragan, Chen, Gupta, Velten, and Gu]{gutierrez2021itof2dtof}
Felipe Gutierrez-Barragan, Huaijin Chen, Mohit Gupta, Andreas Velten, and Jinwei Gu.
\newblock itof2dtof: A robust and flexible representation for data-driven time-of-flight imaging.
\newblock \emph{IEEE Transactions on Computational Imaging}, 7:\penalty0 1205--1214, 2021.

\bibitem[Hu et~al.(2013)Hu, Li, Cheung, and Au]{hu2013depth}
Wei Hu, Xin Li, Gene Cheung, and Oscar Au.
\newblock Depth map denoising using graph-based transform and group sparsity.
\newblock In \emph{2013 IEEE 15th international workshop on multimedia signal processing (MMSP)}, pages 001--006. IEEE, 2013.

\bibitem[Jia et~al.(2025)Jia, He, Wang, Cheung, and Zeng]{jia2025deep}
Jingwei Jia, Changyong He, Jianhui Wang, Gene Cheung, and Jin Zeng.
\newblock Deep unrolled graph laplacian regularization for robust time-of-flight depth denoising.
\newblock \emph{IEEE Signal Processing Letters}, 32:\penalty0 821--825, 2025.

\bibitem[Kang et~al.(2021)Kang, Lee, Jang, and Lee]{kang2021gradient}
Jiwoo Kang, Seongmin Lee, Mingyu Jang, and Sanghoon Lee.
\newblock Gradient flow evolution for 3d fusion from a single depth sensor.
\newblock \emph{IEEE Transactions on Circuits and Systems for Video Technology}, 32\penalty0 (4):\penalty0 2211--2225, 2021.

\bibitem[Kurillo et~al.(2022)Kurillo, Hemingway, Cheng, and Cheng]{kurillo2022evaluating}
Gregorij Kurillo, Evan Hemingway, Mu-Lin Cheng, and Louis Cheng.
\newblock Evaluating the accuracy of the {A}zure {K}inect and {K}inect v2.
\newblock \emph{Sensors}, 22\penalty0 (7):\penalty0 2469, 2022.

\bibitem[Li et~al.(2023)Li, Ye, Wang, Creighton, Taylor, Venkatesh, and Unberath]{li2023temporally}
Zhaoshuo Li, Wei Ye, Dilin Wang, Francis~X Creighton, Russell~H Taylor, Ganesh Venkatesh, and Mathias Unberath.
\newblock Temporally consistent online depth estimation in dynamic scenes.
\newblock In \emph{Proceedings of the IEEE/CVF winter conference on applications of computer vision}, pages 3018--3027, 2023.

\bibitem[Marco et~al.(2017)Marco, Hernandez, Munoz, Dong, Jarabo, Kim, Tong, and Gutierrez]{marco2017deeptof}
Julio Marco, Quercus Hernandez, Adolfo Munoz, Yue Dong, Adrian Jarabo, Min~H Kim, Xin Tong, and Diego Gutierrez.
\newblock Deeptof: off-the-shelf real-time correction of multipath interference in time-of-flight imaging.
\newblock \emph{ACM Transactions on Graphics (ToG)}, 36\penalty0 (6):\penalty0 1--12, 2017.

\bibitem[Miki et~al.(2022)Miki, Lee, Hwangbo, Wellhausen, Koltun, and Hutter]{miki2022learning}
Takahiro Miki, Joonho Lee, Jemin Hwangbo, Lorenz Wellhausen, Vladlen Koltun, and Marco Hutter.
\newblock Learning robust perceptive locomotion for quadrupedal robots in the wild.
\newblock \emph{Science Robotics}, 7\penalty0 (62):\penalty0 eabk2822, 2022.

\bibitem[Min et~al.(2011)Min, Lu, and Do]{min2011depth}
Dongbo Min, Jiangbo Lu, and Minh~N Do.
\newblock Depth video enhancement based on weighted mode filtering.
\newblock \emph{IEEE Transactions on Image Processing}, 21\penalty0 (3):\penalty0 1176--1190, 2011.

\bibitem[Monga et~al.(2021)Monga, Li, and Eldar]{monga2021algorithm}
Vishal Monga, Yuelong Li, and Yonina~C Eldar.
\newblock Algorithm unrolling: Interpretable, efficient deep learning for signal and image processing.
\newblock \emph{IEEE Signal Processing Magazine}, 38\penalty0 (2):\penalty0 18--44, 2021.

\bibitem[Ortega et~al.(2018)Ortega, Frossard, Kova{\v{c}}evi{\'c}, Moura, and Vandergheynst]{ortega2018graph}
Antonio Ortega, Pascal Frossard, Jelena Kova{\v{c}}evi{\'c}, Jos{\'e}~MF Moura, and Pierre Vandergheynst.
\newblock Graph signal processing: Overview, challenges, and applications.
\newblock \emph{Proceedings of the IEEE}, 106\penalty0 (5):\penalty0 808--828, 2018.

\bibitem[Pang and Cheung(2017)]{pang2017graph}
Jiahao Pang and Gene Cheung.
\newblock Graph laplacian regularization for image denoising: Analysis in the continuous domain.
\newblock \emph{IEEE Transactions on Image Processing}, 26\penalty0 (4):\penalty0 1770--1785, 2017.

\bibitem[Pang and Zeng(2021)]{pang2021graph}
Jiahao Pang and Jin Zeng.
\newblock Graph spectral image restoration.
\newblock \emph{Graph Spectral Image Processing}, 133, 2021.

\bibitem[Paszke et~al.(2017)Paszke, Gross, Chintala, Chanan, Yang, DeVito, Lin, Desmaison, Antiga, and Lerer]{paszke2017automatic}
Adam Paszke, Sam Gross, Soumith Chintala, Gregory Chanan, Edward Yang, Zachary DeVito, Zeming Lin, Alban Desmaison, Luca Antiga, and Adam Lerer.
\newblock Automatic differentiation in pytorch.
\newblock 2017.

\bibitem[Patil et~al.(2020)Patil, Van~Gansbeke, Dai, and Van~Gool]{patil2020don}
Vaishakh Patil, Wouter Van~Gansbeke, Dengxin Dai, and Luc Van~Gool.
\newblock Don’t forget the past: Recurrent depth estimation from monocular video.
\newblock \emph{IEEE Robotics and Automation Letters}, 5\penalty0 (4):\penalty0 6813--6820, 2020.

\bibitem[Qiao et~al.(2022)Qiao, Ge, Deng, Wei, Poggi, and Mattoccia]{qiao2022depth}
Xin Qiao, Chenyang Ge, Pengchao Deng, Hao Wei, Matteo Poggi, and Stefano Mattoccia.
\newblock Depth restoration in under-display time-of-flight imaging.
\newblock \emph{IEEE Transactions on Pattern Analysis and Machine Intelligence}, 45\penalty0 (5):\penalty0 5668--5683, 2022.

\bibitem[Ronneberger et~al.(2015)Ronneberger, Fischer, and Brox]{ronneberger2015u}
Olaf Ronneberger, Philipp Fischer, and Thomas Brox.
\newblock U-{N}et: {C}onvolutional networks for biomedical image segmentation.
\newblock In \emph{International Conference on Medical Image Computing and Computer-Assisted Intervention}, pages 234--241. Springer, 2015.

\bibitem[Rossi et~al.(2020)Rossi, Gheche, Kuhn, and Frossard]{rossi2020joint}
Mattia Rossi, Mireille~El Gheche, Andreas Kuhn, and Pascal Frossard.
\newblock Joint graph-based depth refinement and normal estimation.
\newblock In \emph{Proceedings of the IEEE/CVF Conference on Computer Vision and Pattern Recognition}, pages 12154--12163, 2020.

\bibitem[Schelling et~al.(2022)Schelling, Hermosilla, and Ropinski]{schelling2022radu}
Michael Schelling, Pedro Hermosilla, and Timo Ropinski.
\newblock Radu: Ray-aligned depth update convolutions for tof data denoising.
\newblock In \emph{Proceedings of the IEEE/CVF Conference on Computer Vision and Pattern Recognition}, pages 671--680, 2022.

\bibitem[Su et~al.(2018)Su, Heide, Wetzstein, and Heidrich]{su2018deep}
Shuochen Su, Felix Heide, Gordon Wetzstein, and Wolfgang Heidrich.
\newblock Deep end-to-end time-of-flight imaging.
\newblock In \emph{Proceedings of the IEEE Conference on Computer Vision and Pattern Recognition}, pages 6383--6392, 2018.

\bibitem[Sun et~al.(2023)Sun, Ye, Xiong, Choe, Wang, Su, and Ranjan]{sun2023consistent}
Zhanghao Sun, Wei Ye, Jinhui Xiong, Gyeongmin Choe, Jialiang Wang, Shuochen Su, and Rakesh Ranjan.
\newblock Consistent direct time-of-flight video depth super-resolution.
\newblock In \emph{Proceedings of the IEEE/CVF Conference on Computer Vision and Pattern Recognition}, pages 5075--5085, 2023.

\bibitem[Xiang et~al.(2016)Xiang, Echtler, Kerl, Wiedemeyer, Gordon, and Facioni]{xiang2016libfreenect2}
Lingzhu Xiang, Florian Echtler, Christian Kerl, Thiemo Wiedemeyer, R Gordon, and F Facioni.
\newblock libfreenect2: Release 0.2, 2016.

\bibitem[Yan et~al.(2018)Yan, Wu, Wang, Xu, An, Guo, and Liu]{yan2018ddrnet}
Shi Yan, Chenglei Wu, Lizhen Wang, Feng Xu, Liang An, Kaiwen Guo, and Yebin Liu.
\newblock Ddrnet: Depth map denoising and refinement for consumer depth cameras using cascaded cnns.
\newblock In \emph{Proceedings of the European conference on computer vision (ECCV)}, pages 151--167, 2018.

\bibitem[Yang et~al.(2014)Yang, Mao, Cheung, Stankovic, and Chan]{yang2014graph}
Cheng Yang, Yu Mao, Gene Cheung, Vladimir Stankovic, and Kevin Chan.
\newblock Graph-based depth video denoising and event detection for sleep monitoring.
\newblock In \emph{2014 IEEE 16th international workshop on multimedia signal processing (MMSP)}, pages 1--6. IEEE, 2014.

\bibitem[Zanuttigh et~al.(2016)Zanuttigh, Marin, Dal~Mutto, Dominio, Minto, Cortelazzo, et~al.]{zanuttigh2016time}
Pietro Zanuttigh, Giulio Marin, Carlo Dal~Mutto, Fabio Dominio, Ludovico Minto, Guido~Maria Cortelazzo, et~al.
\newblock Time-of-flight and structured light depth cameras.
\newblock \emph{Technology and Applications}, 978\penalty0 (3), 2016.

\bibitem[Zeng et~al.(2019)Zeng, Pang, Sun, and Cheung]{zeng2019deep}
Jin Zeng, Jiahao Pang, Wenxiu Sun, and Gene Cheung.
\newblock Deep graph laplacian regularization for robust denoising of real images.
\newblock In \emph{Proceedings of the ieee/cvf conference on computer vision and pattern recognition workshops}, pages 0--0, 2019.

\bibitem[Zhang et~al.(2022)Zhang, Cheung, Pang, Sanghvi, Gnanasambandam, and Chan]{zhang2022graph}
Xue Zhang, Gene Cheung, Jiahao Pang, Yash Sanghvi, Abhiram Gnanasambandam, and Stanley~H Chan.
\newblock Graph-based depth denoising \& dequantization for point cloud enhancement.
\newblock \emph{IEEE Transactions on Image Processing}, 31:\penalty0 6863--6878, 2022.

\end{thebibliography}
}

\clearpage
\setcounter{page}{1}
\maketitlesupplementary

In this supplementary material, we provide the derivation of the data fidelity term in MAP formulation based on ToF depth noise distribution in Sec.\,\ref{sec:likelihood}.
Then we evaluate the sensitivity to frame time step in Sec.\,\ref{sec:framegap}.
Next, we summarize unrolling of cross-frame graph fusion based
ToF depth denoising algorithm in Sec.\,\ref{sec:al_sum}.
More visualization results are provided in Sec.\,\ref{sec:vis}, with a video demonstrating the estimation accuracy and temporal consistency.

\section{Data Fidelity Term in MAP Problem}
\label{sec:likelihood}

In this section, we derive the data fidelity term based on ToF depth noise distribution.
As assumed in Sec.\,\ref{sec:3_map}, $\x_i$ and $\x_q$ are corrupted by additive white Gaussian noise (AWGN) \cite{frank2009theoretical,georgiev2018time}, and the pixels in $\y_i^t,\y_q^t$ are independent and identically distributed with multivariate Gaussian distribution.
The joint probability density function of $\y_i^t,\y_q^t$ is given as:
\begin{multline}\label{eq:pdf_iq}
P(\y_i^t,\y_q^t|\x_i^t,\x_q^t) = 
\frac{1}{(2\pi\sigma^2)^N} \\
\times \exp\left(
 -\frac{ (\n_i^t)^{\top}\n_i^t + (\n_q^t)^{\top}\n_q^t }{2\sigma^2}
\right),
\end{multline}
\vspace{-2ex}
\begin{align}
\n_i^t &= \y_i^t - \x_i^t, \quad
\n_q^t = \y_q^t - \x_q^t,
\end{align}

where $\sigma$ is the noise variance.

Since the final target is to reconstruct depth, we further investigate depth noise distribution based on (\ref{eq:pdf_iq}).
Based on (\ref{eq:d}) and (\ref{eq:pdf_iq}), the distribution of depth noise $\n^t_d$ is derived in \cite{frank2009theoretical,georgiev2018time} as,
\begin{align}  \nonumber
    &P(\n^t_d) = \\ \nonumber
    & \prod_{m=1}^N \frac{\cos (4\pi f_m \n^t_d(m)/c)}{2\gamma^t(m) \sqrt{2\pi}} [ \, 1+ \mathrm{erf}( \, \frac{\cos (4\pi f_m \n^t_d(m)/c)}{\gamma^t(m)\sqrt{2}} ) \, ] \, \\ \label{eq:pdf_d_s}
    &  \times \exp ( \, -\frac{\sin^2(4\pi f_m \n^t_d(m)/c)}{2\gamma^t(m)^2} ) \, + \frac{1}{2\pi} \exp(-\frac{1}{2\gamma^t(m)^2}),
\end{align} 
where $\gamma^t(m) = \sigma/\y^t_a(m)$, $\y^t_a(m)$ is noisy amplitude, $\mathrm{erf}$ is the Gaussian error function. 
Under normal noise level, \ie, $\gamma^t(m) \ll 1$, we have $\mathrm{erf} $ output equal to $1$ and last term equal to $0$ in (\ref{eq:pdf_d_s}), then (\ref{eq:pdf_d_s}) is approximated with


\begin{align} \nonumber
P(\n^t_d) \approx \prod_{m=1}^N  ( &\frac{\cos (4\pi f_m \n^t_d(m)/c)}{\gamma^t(m) \sqrt{2\pi}} \\ \label{eq:pdf_d_simple}
& \times \exp ( \, -\frac{\sin^2(4\pi f_m \n^t_d(m)/c)}{2(\gamma^t(m))^2} ) ).
\end{align} 

Based on (\ref{eq:pdf_d_simple}), the log of likelihood $P(\y^t_d|\x^t_d)$ is 
\begin{align} \nonumber
 \ln P(\y^t_d|\x^t_d) &\approx \sum_{m=1}^N (\ln (\cos (4\pi f_m \n_d^t(m)/c)) - \\  \label{eq:pdf_d_log}
 &\sin^2(4\pi f_m \n_d^t(m)/c)/(2\gamma^t(m)^2)),
\end{align} 
where the irrelevant term 
$-\ln(\gamma^t(m) \sqrt{2\pi}) $ is removed.
Both terms in (\ref{eq:pdf_d_log}) minimize $n_d$, and with $\gamma \ll 1$, the second term dominates.
Thus, we remove the first term and compute the likelihood as a function of $\x^t_i,\x^t_q$ as follows:
\begin{multline}\label{eq:pdf_d_simple2}
\ln P(\y^t_d|\x^t_d) \approx 
\sum_{m=1}^N -\frac{ \sin^2 \bigl( \phi(m) - \phi'(m) \bigr) }{ 2 \gamma^t(m)^2 } \\
= \sum_{m=1}^N -\frac{\bigl( \sin \phi(m) \cos \phi'(m) - \cos \phi(m) \sin \phi'(m) \bigr)^2}{2 \gamma^t(m)^2},
\end{multline}

where $\phi'= 4\pi f_m \y^t_d/c$ is the noisy phase. 


Based on (\ref{eq:pdf_d_simple2}) and (\ref{eq:d}), the log of likelihood of $\n^t_d$ is given as a function of $\x_i^t,\x_q^t$:
\begin{equation} \label{eq:likely2}
\ln P(\n^t_d) \approx  -\frac{1}{2\sigma^2} \| (\X_a^t)^{-1}(\x^t_q \odot \y^t_i - \x^t_i \odot \y^t_q) \|_2^2,
\end{equation}
where $\X_a^t = \mathrm{diag}(\x^t_a)$ is the amplitude, $\odot$ is Hadamard product.


\section{Analysis of Frame Time Step}
\label{sec:framegap}

\begin{table}[b]
\centering
\small
\caption{Comparison of quantitative evaluation on DvToF testing dataset with different frame rate}
\label{tab:ablation}
\begin{tabular}{ccccc}
\hline
Time step & MAE(m)$\downarrow$ & AbsRel$\downarrow$ & $\delta_1$$\uparrow$ & TEPE(m) $\downarrow$  \\
\hline
1 &  \textbf{0.0190}&\textbf{0.0060}& \textbf{0.9974} &0.0634\\
2 &  0.0192 &\textbf{0.0060}& 0.9973& \textbf{0.0608}  \\
4 & 0.0194 &0.0062& 0.9972 & 0.0647 \\
8 &  0.0210 &0.0071 &0.9970 &0.0725\\
\hline
\end{tabular}
\end{table}

Following \cite{dong2024exploiting}, to investigate the effect of time step between reference and current frames,
we test GIGA-ToF on DVToF dataset with different time steps.
For small time steps $\Delta t = 1,2$, the performances are similar.
When time steps become larger, $\Delta t = 4,8$ reduces due to the limited similarity of graph structures in the neighboring pixels in the reference frame.
Nevertheless, the performance still surpasses that of single-frame processing, validating the necessity of multi-frame processing and the temporal self-similarity of graph structures despite large frame gaps.

\section{Algorithm Summary}
\label{sec:al_sum}

Based on the algorithm unrolling of graph Laplacian regularization, we obtain the solution to (\ref{eq:local_op}), which is summarized in Algorithm \ref{algo:giga}.

\begin{algorithm}[h]
\caption{Unrolling of Cross-frame Graph Fusion based ToF Depth Denoising Algorithm}
\label{algo:giga}
\begin{small}
\begin{algorithmic}[1]
\Require Noisy ToF raw data $\y^t_i,\y^t_q$, intra-frame graph adjacency matrices $\W^t_i, \W^t_q, \W^{t-1}_i, \W^{t-1}_q$, inter-frame graph adjacency matrix $\W^{t,t-1}$ and fusion weight $\bPhi^{t,t-1}$, GLR prior weight $\bLambda_i^t,\bLambda_q^t$, iteration number $R,T$
\Ensure Denoised output $\x^t_i,\x^t_q$
\State Map reference frame graphs $\W^{t-1}_i, \W^{t-1}_q$ to current frame to obtain mapped graphs $\hat{\W}^{t-1}_i, \hat{\W}^{t-1}_q$ using (\ref{eq:w_map})
\State Fuse the mapped graphs with current frame graphs to obtain fused graphs $\widetilde{\W}^t_i, \widetilde{\W}^t_q $ using (\ref{eq:w_fuse}) 
\State Obtain corresponding $\widetilde{\D}^t_i, \widetilde{\D}^t_q$ from $\widetilde{\W}^t_i , \widetilde{\W}^t_q $ using (\ref{eq:deg})
\State Initialize $\x_i^{t,0} = \y_i^t,\x_q^{t,0} = \y_q^t$
\For{$r$ = $0:R-1$}
    \State Update $\bLambda_i^{t,r-1}$ with $\X_a^{t,r-1}$, fix $\x_q^{t,r}$ and optimize $\x_i^{t,r}$
    \For{$p$ = $0:P-1$} \label{al:p}
         \State Transform $\x_i^{t,r,p}$ with convolutional kernel $\widetilde{\W}_i^t$ 
         \State Fuse with $\x_i^{t,r-1}$ with weight $\bLambda_i^{t,r-1}$ as specified in (\ref{eq:local_op}) to update $\mathbf{x}_i^{t,r,p+1}$
    \EndFor \label{al:p1}
    \State Fix $\x_i^{t,r}$ and repeat steps \ref{al:p}-\ref{al:p1} to optimize $\x_q^{t,r}$
\EndFor
\State Output 
\end{algorithmic}
\end{small}
\end{algorithm}


\section{More Visualization}
\label{sec:vis}

\begin{figure}[t]
\centering
\includegraphics[width=\columnwidth]{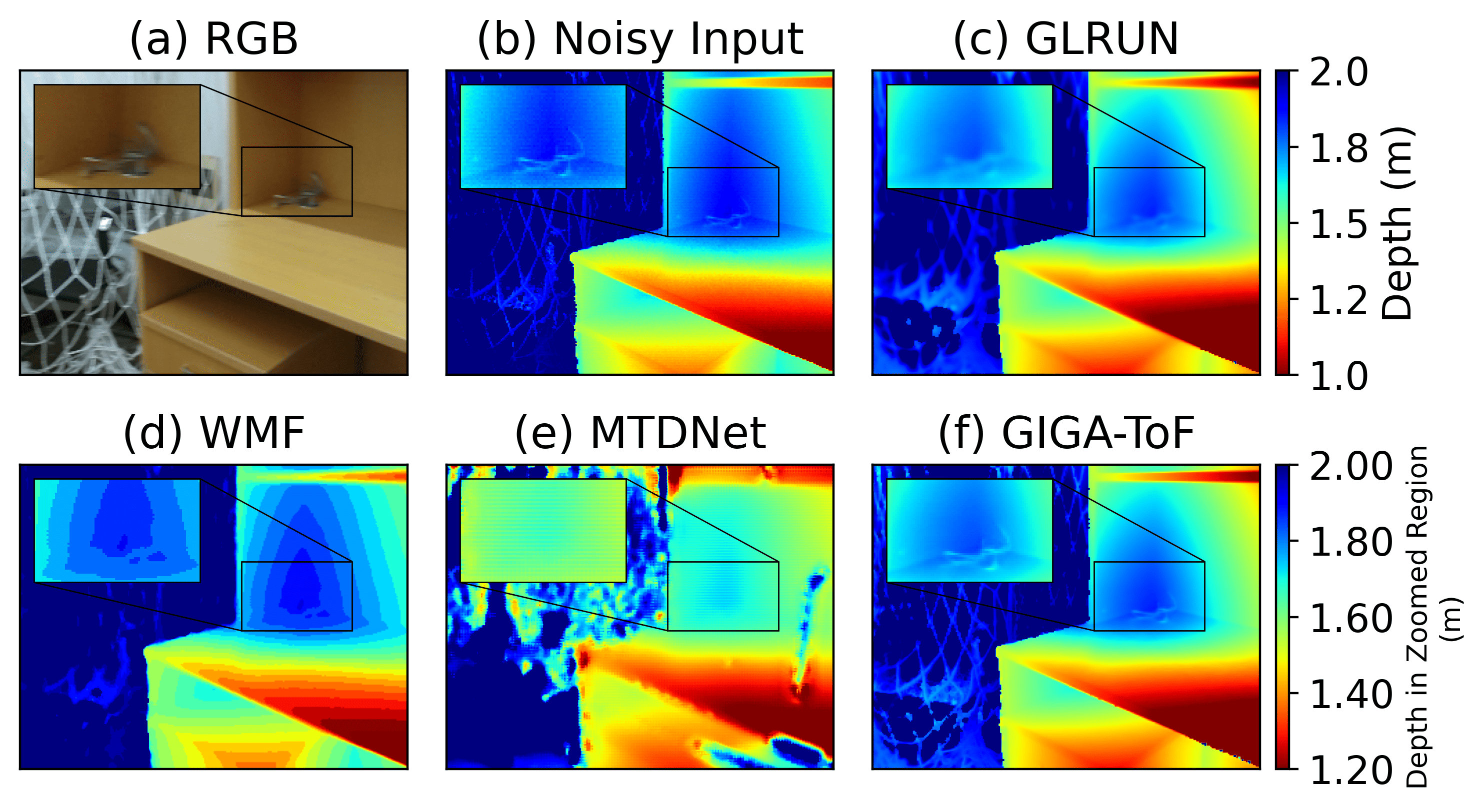}
\caption{Visual results of ToF depth denoising on real data captured by Kinect v2 sensor: (a) RGB and (b) noisy depth captured by Kinectv2 camera, and results of (c) GLRUN, (d) WMF, (e) MTDNet and (f) GIGA-ToF, where GIGA-ToF shows accurate and smooth estimation.}
\label{fig:kinect-supp1}
\end{figure}

\begin{figure}[t]
\centering
\includegraphics[width=\columnwidth]{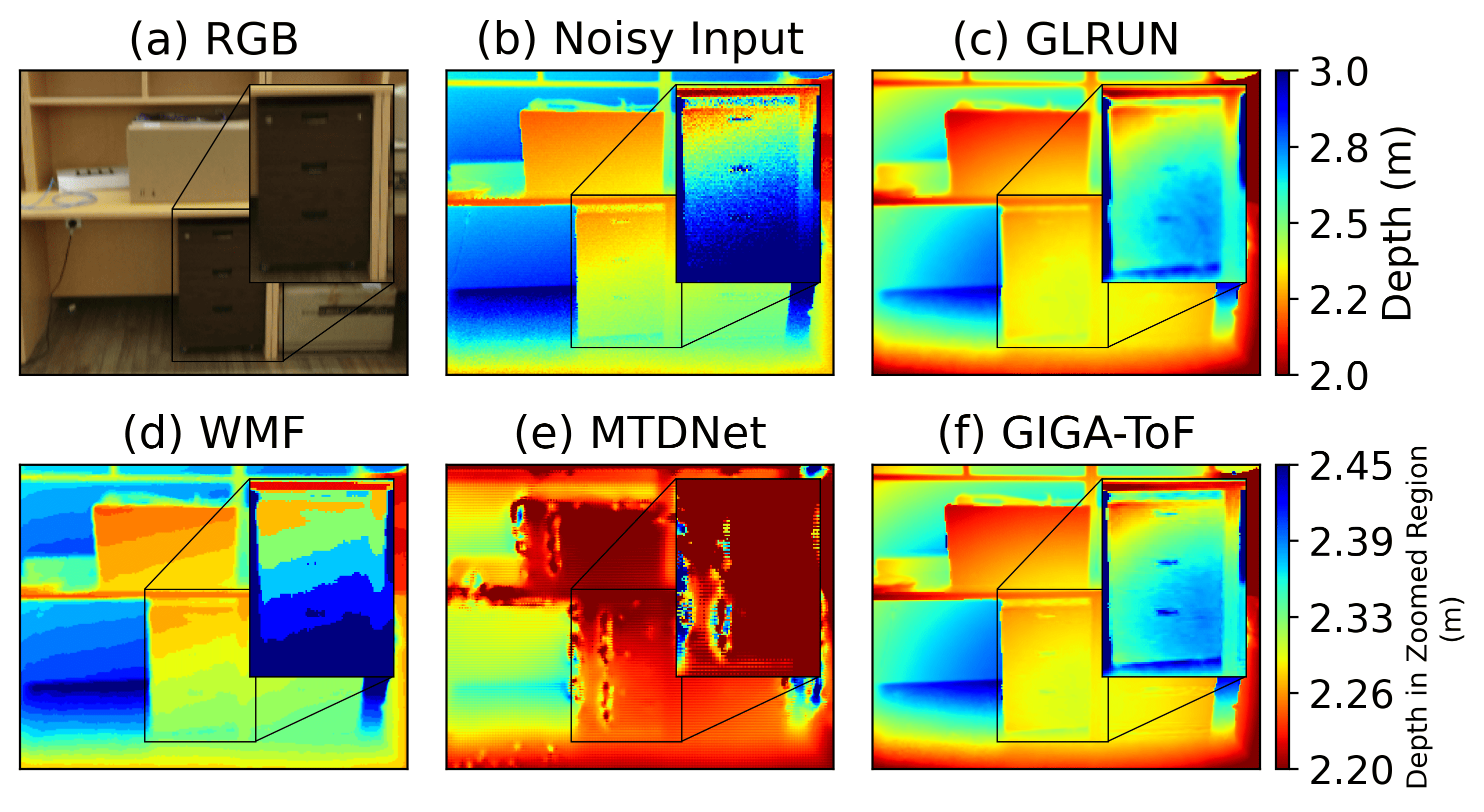}
\caption{Visual results of ToF depth denoising on real data captured by Kinect v2 sensor: (a) RGB and (b) noisy depth captured by Kinectv2 camera, and results of (c) GLRUN, (d) WMF, (e) MTDNet and (f) GIGA-ToF, where GIGA-ToF shows robustness to real noise and recovers accurate details.}
\label{fig:kinect-supp2}
\end{figure}

\begin{figure}[t]
\centering
\includegraphics[width=\columnwidth]{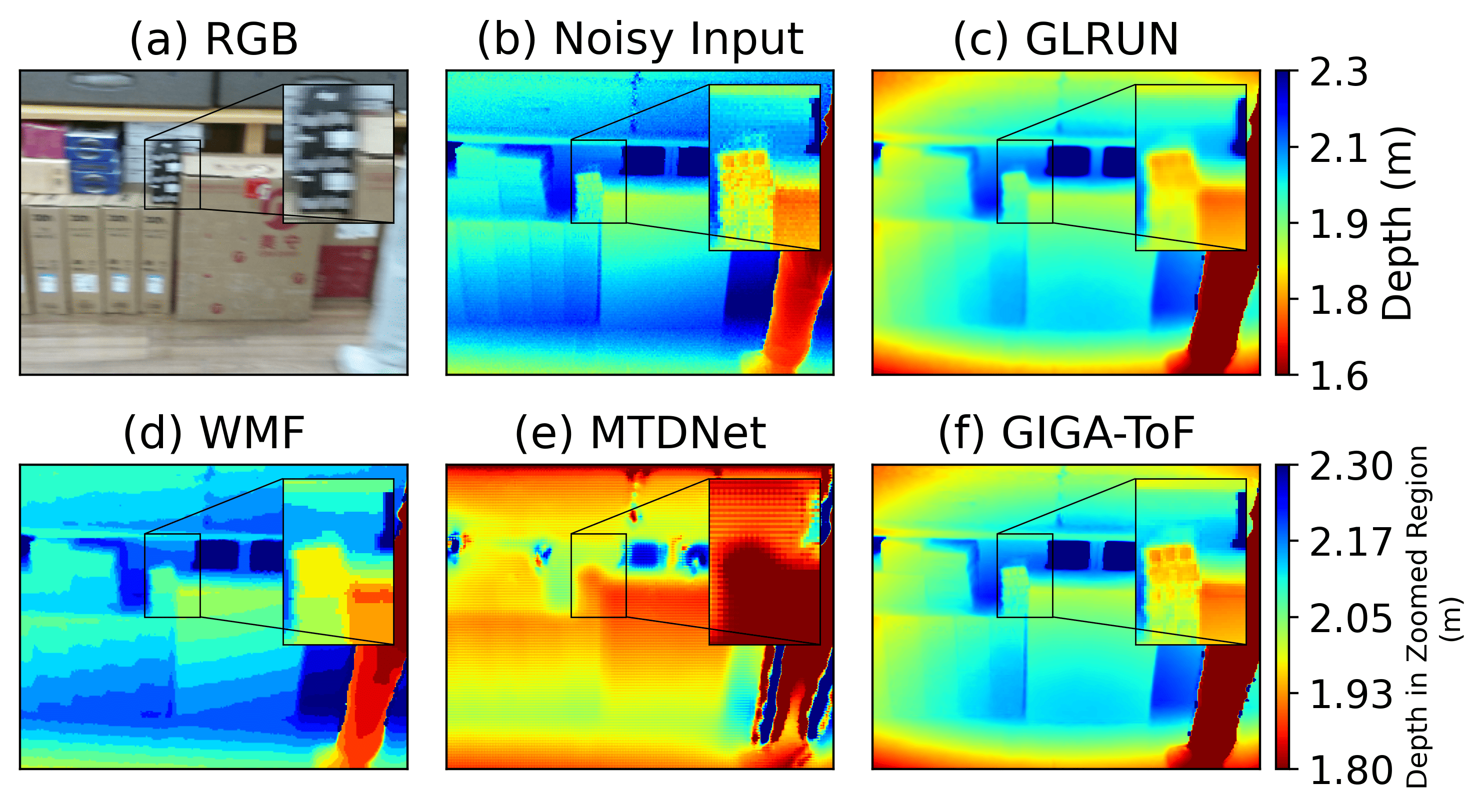}
\caption{Visual results of ToF depth denoising on real data captured by Kinect v2 sensor: (a) RGB and (b) noisy depth captured by Kinectv2 camera, and results of (c) GLRUN, (d) WMF, (e) MTDNet and (f) GIGA-ToF, where GIGA-ToF exhibits better spatial sharpness than other competing schemes.}
\label{fig:kinect-supp3}
\end{figure}

\begin{figure*}[ht]
\centering
\includegraphics[width=\textwidth]{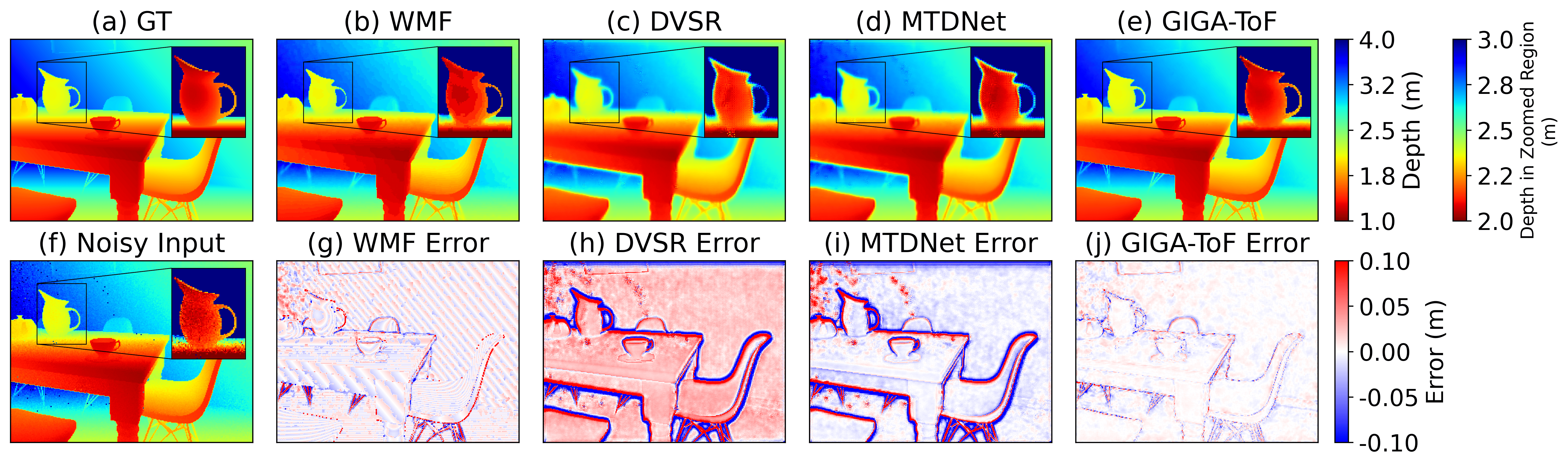}
\caption{Depth results and error maps of ToF depth denoised on DvToF dataset: (a) GT, results of (b) WMF \cite{min2011depth}, (c) DVSR \cite{sun2023consistent}, (d) MTDNet \cite{dong2024exploiting} and (e) proposed GIGA-ToF. Corresponding error maps are in the second row.
GIGA-ToF shows more accurate depth estimation, maintaining global smoothness with edge preservation, \eg, the teapot handle highlighted in the zoom-in block.}
\label{fig:exp_dvtof_supp}
\end{figure*}

\begin{figure*}[t]
\centering
\includegraphics[width=\textwidth]{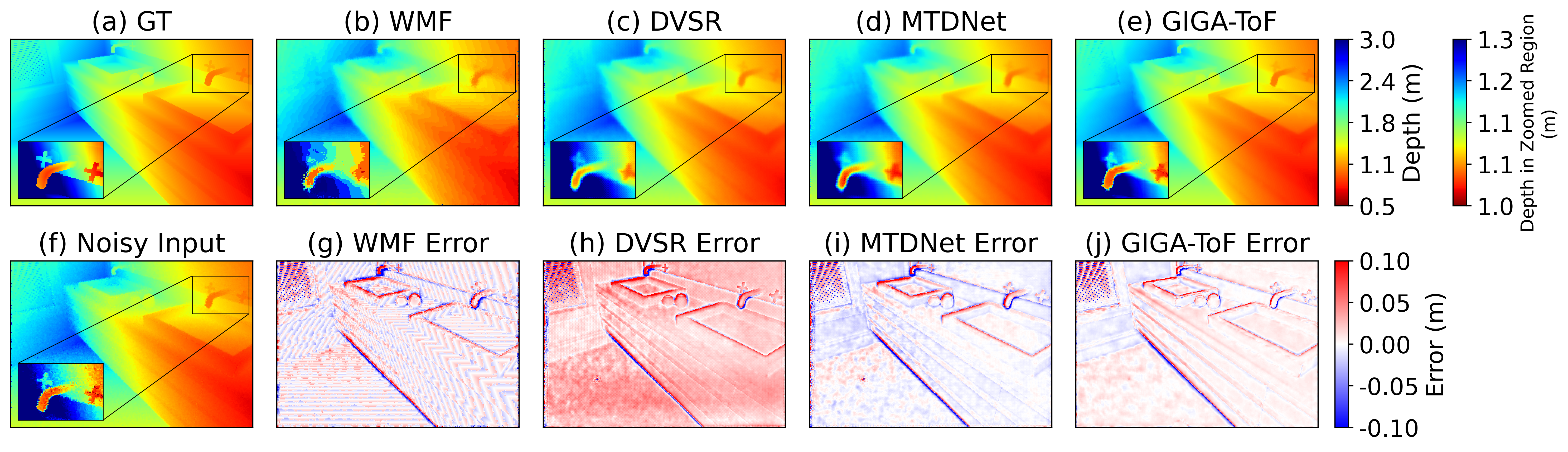}
\caption{Depth results and error maps of ToF depth denoised on DvToF dataset: (a) GT, results of (b) WMF \cite{min2011depth}, (c) DVSR \cite{sun2023consistent}, (d) MTDNet \cite{dong2024exploiting} and (e) proposed GIGA-ToF. Corresponding error maps are in the second row.
While MTDNet shows competing results, the details are blurred as highlighted in the zoom-in block, while GIGA-ToF generates sharp edges due to utilization of motion-invariant graph structure fusion.}
\label{fig:exp_dvtof_barron_supp}
\end{figure*}

\begin{figure*}[t]
\centering
\includegraphics[width=\textwidth]{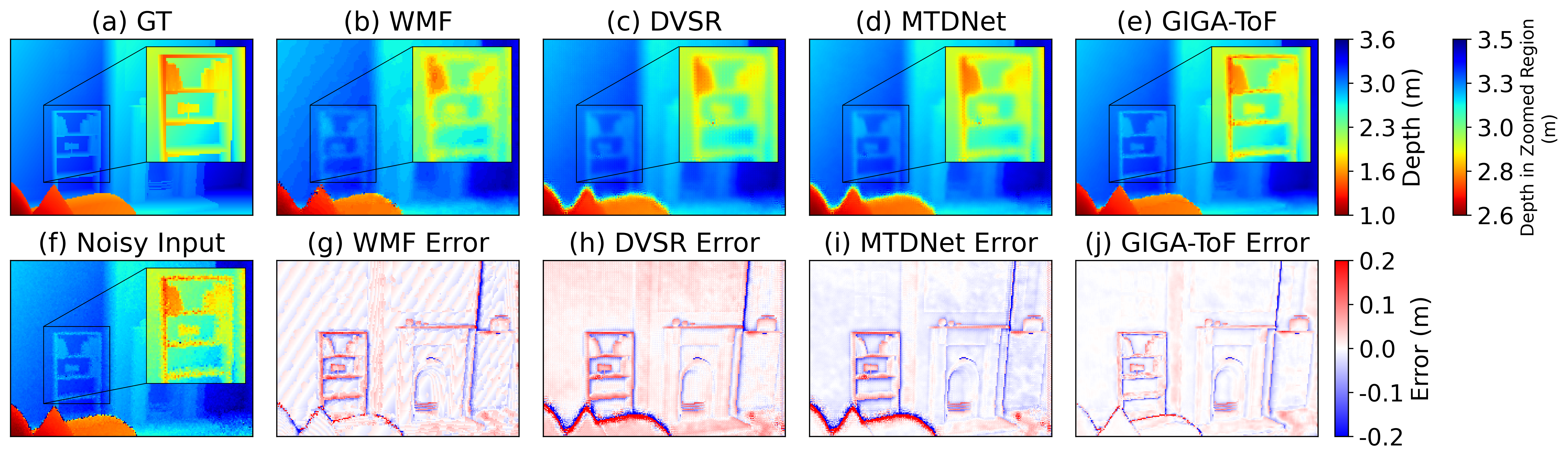}
\caption{Depth results and error maps of ToF depth denoised on DvToF dataset with augmented edge noise: (a) GT, results of (b) WMF \cite{min2011depth}, (c) DVSR \cite{sun2023consistent}, (d) MTDNet \cite{dong2024exploiting} and (e) proposed GIGA-ToF. Corresponding error maps are in the second row. GIGA-ToF shows strong generalization to unseen edge noise and generates accurate details, \eg, in the bookshelf in the zoom-in block.}
\label{fig:exp_dvtof_barron_supp_2}
\end{figure*}

We provide more results for the qualitative comparison of ToF depth denoising methods.
In particular, we demonstrate results on synthetic DVToF dataset in Fig.~\ref{fig:exp_dvtof_supp} and Fig.~\ref{fig:exp_dvtof_barron_supp}, and DVToF dataset with noise augmentation in Fig.~\ref{fig:exp_dvtof_barron_supp_2}.
To further demonstrate the generalization ability to real Kinectv2 data, we shown results in Figs.~\ref{fig:kinect-supp1}, \ref{fig:kinect-supp2}, \ref{fig:kinect-supp3}.
Note that we directly apply the model trained on original DVToF dataset to the noise-augmented DVToF dataset and Kinectv2 dataset without fine-tuning, which validates its generalization ability.
Please kindly refer to the supplementary video for better temporal visualizations.

\end{document}